\pdfoutput=1

\documentclass[11pt]{article}

\usepackage{EMNLP2023}

\usepackage{times}
\usepackage{latexsym}
\usepackage{amsmath}

\usepackage[T1]{fontenc}

\usepackage[utf8]{inputenc}

\usepackage{microtype}

\usepackage{inconsolata}

\usepackage{subcaption}
\usepackage{comment}
\usepackage{todonotes}

\title{Neurons in Large Language Models: Dead, N-gram,  Positional}

\author{Elena Voita$^{1}$ \quad \ Javier Ferrando$^{2\ast}$ \quad \ Christoforos Nalmpantis$^{1}$  \\
 $^{1}$Meta AI\\ $^{2}$TALP Research Center, Universitat Politècnica de Catalunya\\ 
  \texttt{\{lenavoita,
christoforos\}@meta.com}, \texttt{javier.ferrando.monsonis@upc.edu} \\}

\let\svthefootnote\thefootnote

\begin{document}
\maketitle
\begin{abstract}

We analyze a family of large language models in such a lightweight manner that can be done on a single GPU. Specifically, we focus on the OPT family of models ranging from 125m to 66b parameters and rely only on whether an FFN neuron is activated or not. First, we find that the early part of the network is sparse and represents many discrete features. Here, many neurons (more than $70\%$ in some layers of the 66b model) are ``dead'', i.e. they never activate on a large collection of diverse data. At the same time, many of the alive neurons are reserved for discrete features and act as token and n-gram detectors. Interestingly, their corresponding FFN updates not only promote next token candidates as could be expected, but also explicitly focus on removing the information about triggering them tokens, i.e., current input. To the best of our knowledge, this is the first example of mechanisms specialized at removing (rather than adding) information from the residual stream. With scale, models become more sparse in a sense that they have more dead neurons and token detectors. Finally, some neurons are positional: them being activated or not depends largely (or solely) on position and less so (or not at all) on textual data. We find that smaller models have sets of neurons acting as position range indicators while larger models operate in a less explicit manner.

\end{abstract}

\let\thefootnote\relax\footnote{$^{\ast}$Work done as part of internship at Meta AI.}
\addtocounter{footnote}{-1}\let\thefootnote\svthefootnote

\section{Introduction}

The range of capabilities of language models expands with scale and at larger scales models become so strong and versatile that a single model can be integrated into various applications and decision-making processes~\citep{brown-gpt3,kaplan2020scaling,wei2022emergent,ouyang2022training,openai2023gpt4,anil2023palm}. This increases interest and importance of understanding the internal workings of these large language models (LLMs) and, specifically, their evolution with scale. 
Unfortunately, scaling also increases the entry threshold for interpretability researchers since dealing with large models requires a lot of computational resources. In this work, we analyze a family of OPT models up to 66b parameters and deliberately keep our analysis very lightweight so that it could be done using a single GPU.

We focus on neurons inside FFNs, i.e. individual activations in the representation between the two linear layers of the Transformer feedforward blocks (FFNs). Differently from e.g. neurons in the residual stream, FFN neurons are more likely to represent meaningful features: the elementwise nonlinearity breaks the rotational invariance of this representation and encourages features to align with the basis dimensions~\citep{elhage2021mathematical}. When such a neuron is activated, it updates the residual stream by pulling out the corresponding row of the second FFN layer; when it is not activated, it does not update the residual stream (Figure~\ref{fig:suppressed_concepts}).\footnote{Since OPT models have the ReLU activation function, the notion of ``activated'' or ``not activated'' is trivial and means non-zero vs zero.} Therefore, we can interpret functions of these FFN neurons in two ways: (i) by understanding when they are activated, and (ii) by interpreting the corresponding updates coming to the residual stream.

First, we find that in the first half of the network, 
many neurons are ``dead'', i.e. they never activate on a large collection of diverse data. Larger models are more sparse in this sense: for example, in the 66b model more that $70\%$ of the neurons in some layers are dead. At the same time, many of the alive neurons in this early part of the network are reserved for discrete features and act as indicator functions for tokens and n-grams: they activate if and only if the input is a certain token or an n-gram. The function of the updates coming from these token detectors to the residual stream is also very surprising: at the same time as they promote concepts related to the potential next token candidate (which is to be expected according to~\citet{geva-etal-2021-transformer,geva-etal-2022-transformer}), they are \textit{explicitly targeted at removing information about current input}, i.e. their triggers. This means that in the bottom-up processing where a representation of the current input token gets gradually transformed into a representation for the next token, current token identity is removed by the model explicitly (rather than ends up implicitly ``buried'' as a result of additive updates useful for the next token). To the best of our knowledge, this is the first example of mechanisms specialized at removing (rather than adding) information from the residual stream. 

Finally, we find that some neurons are responsible for encoding positional information regardless of textual patterns. Similarly to token and n-gram detectors, many of these neurons act as indicator functions of position ranges, i.e. activate for positions within certain ranges and do not activate otherwise. Interestingly, these neurons often collaborate. For example, the second layer of the 125m model has 10 positional neurons whose indicated positional ranges are in agreement: together, they efficiently cover all possible positions and no neuron is redundant. In a broader picture, positional neurons question the key-value memory view of the FFN layers stating that ``each key correlates with textual patterns in the training data and each value induces a distribution over the output vocabulary''~\cite{geva-etal-2021-transformer,geva-etal-2022-transformer}. Neurons that rely on position regardless of textual pattern indicate that FFN layers can be used by the model in ways that \textit{do not fit the key-value memory view}. Overall, we argue that the roles played by these layers are still poorly understood.

Overall, we find neurons that:
\begin{itemize}
    \item are ``dead'', i.e. never activate on a large diverse collection of data;
    \item act as token- and n-gram detectors that, in addition to promoting next token candidates, explicitly remove current token information;
    \item encode position regardless of textual content which indicates that the role of FFN layers extends beyond the key-value memory view.
\end{itemize}

With scale, models have more dead neurons and token detectors and are less focused on absolute position.

\section{Data and Setting}

\paragraph{Models.} We use OPT~\cite{zhang2022opt}, a suite of decoder-only pre-trained transformers that are publicly available. We use model sizes ranging from 125M to 66B parameters and take model weights from the HuggingFace model hub.\footnote{\url{https://huggingface.co/models}}

\paragraph{Data.} We use data from diverse sources containing development splits of the datasets used in OPT training as well as several additional datasets. Overall, we used (i)~subsets of the validation and test part of the Pile~\cite{gao2020pile} including Wikipedia, DM Mathematics, HackerNews,  (ii)~Reddit\footnote{Pushshift.io Reddit dataset is a previously existing dataset extracted and obtained by a third party that contains preprocessed comments posted on the social network Reddit and hosted by pushshift.io.}~\cite{baumgartner2020pushshift,roller-etal-2021-recipes}, (iii)~code data from Codeparrot\footnote{\url{https://huggingface.co/datasets/codeparrot/codeparrot-clean}}. 

For the experiments in Section~\ref{sect:dead_neurons} when talking about dead neurons, we use several times more data. Specifically, we add more data from Wikipedia, DM Mathematics and Codeparrot, as well as add new domains from the Pile\footnote{\url{https://huggingface.co/datasets/EleutherAI/pile}}: EuroParl, FreeLaw, PubMed abstracts, Stackexchange.

Overall, the data used in Section~\ref{sect:dead_neurons} has over 20M tokens, in the rest of the paper~-- over 5M tokens.

\paragraph{Single-GPU processing.} We use only sets of neuron values for some data, i.e. we run only forward passes of the full model or its several first layers. Since large models do not fit in a single GPU, we load one layer at a time keeping the rest of the layers on CPU. This allows us to record neuron activations for large models: all the main experiments in this paper were done on a single GPU.

\section{Dead Neurons}
\label{sect:dead_neurons}

\begin{figure}[t!]
    \centering
    \begin{subfigure}[b]{0.23\textwidth}
        \includegraphics[width=\textwidth]{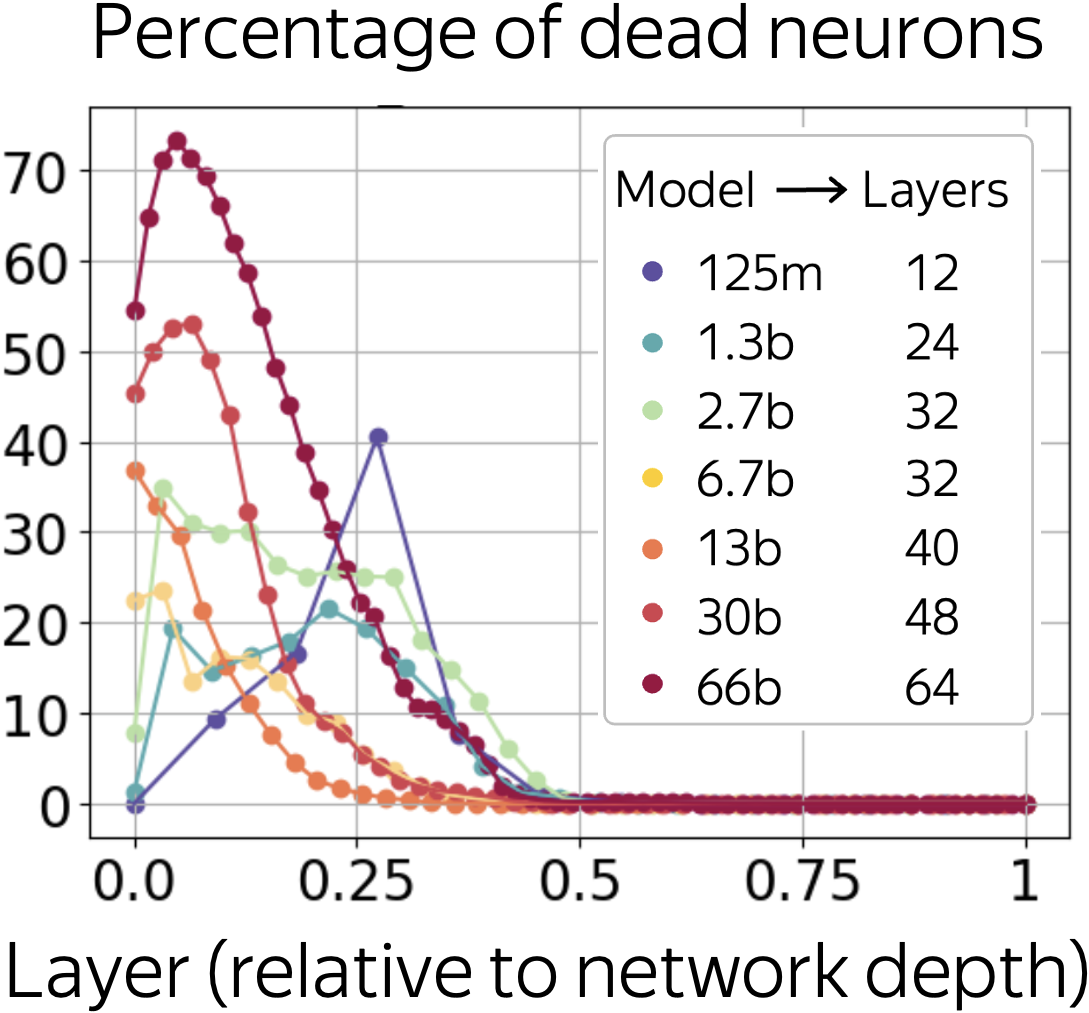}
        \caption{}
        \label{fig:dead_neurons}
    \end{subfigure}
    \ \ 
    \begin{subfigure}[b]{0.23\textwidth}
         \includegraphics[width=\textwidth]{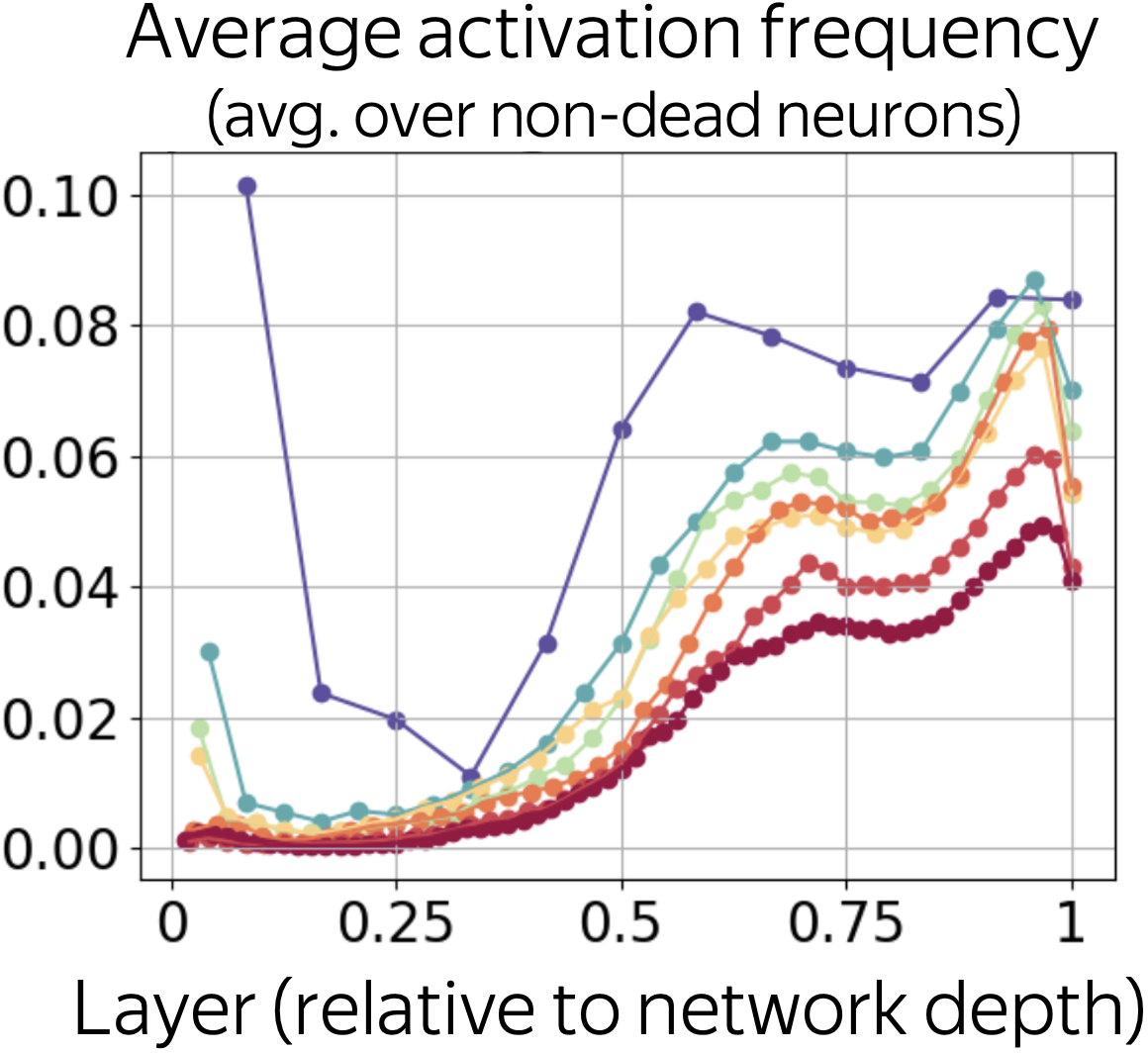}
        \caption{}
        \label{fig:neuron_activation_frequency}
    \end{subfigure}
    \vspace{-2ex}
    \caption{(a) Percentage of ``dead'' neurons; (b) average neuron activation frequency among non-dead neurons.}
    \vspace{-2ex}
    \label{fig:dead_neurons_and_frequency}
\end{figure}

Let us start from simple statistics such as neuron activation frequency (Figure~\ref{fig:dead_neurons_and_frequency}).

\paragraph{Many neurons are ``dead''.}
First, we find that many neurons never activate on our diverse data, i.e. they can be seen as ``dead''. Figure~\ref{fig:dead_neurons}  shows that the proportion of dead neurons is very substantial: e.g., for the 66b model, the proportion of dead neurons in some layers is above $70\%$. We also see that larger models are more sparse because (i) they have more dead neurons and (ii) the ones that are alive activate less frequently (Figure~\ref{fig:neuron_activation_frequency}).

\paragraph{Only first half of the model is sparse.} Next, we notice that this kind of sparsity is specific only to early layers. This leads to a clear distinction between the first and the second halves of the network: while the first half contains a solid proportion of dead neurons, the second half is fully ``alive''. Additionally, layers with most dead neurons are the ones where alive neurons activate most rarely.

\paragraph{Packing concepts into neurons.}
This difference in sparsity across layers might be explained by ``concept-to-neuron'' ratio being much smaller in the early layers than in the higher layers. Intuitively, the model has to represent sets of encoded in a layer concepts by ``spreading'' them across available neurons.
In the early layers, encoded concepts are largely shallow and are likely to be discrete (e.g., lexical) while at the higher layers, networks learn high-level semantics and reasoning~\cite{peters-etal-2018-deep,liu-etal-2019-linguistic,jawahar-etal-2019-bert,tenney-etal-2019-bert,geva-etal-2021-transformer}. Since 
the number of possible shallow patterns is not large and, potentially, enumerable, in the early layers the model can (and, as we will see later, does) assign dedicated neurons to some features. The more neurons are available to the model, the easier it is to do so~-- this agrees with the results in Figure~\ref{fig:dead_neurons_and_frequency} showing that larger models are more sparse. Differently, the space of fine-grained semantic concepts is too large compared to the number of available neurons which makes it hard to reserve many dedicated neuron-concept pairs.\footnote{There can, however, be a few specialized neurons in the higher layers. For example, BERT has neurons responsible for relational facts \cite{dai-etal-2022-knowledge}.}

\paragraph{Are dead neurons completely dead?} Note that the results in Figure~\ref{fig:dead_neurons} can mean one of the two things: (i)~these neurons can never be activated (i.e. they are ``completely dead'') or (ii) they correspond to patterns so rare that we never encountered them in our large diverse collection of data. While the latter is possible, note that this does not change the above discussion about sparsity and types of encoded concepts. On the contrary: it further supports the hypothesis of models assigning dedicated neurons to specific concepts.

\begin{figure}[t]
\centering
{\includegraphics[scale=0.27]{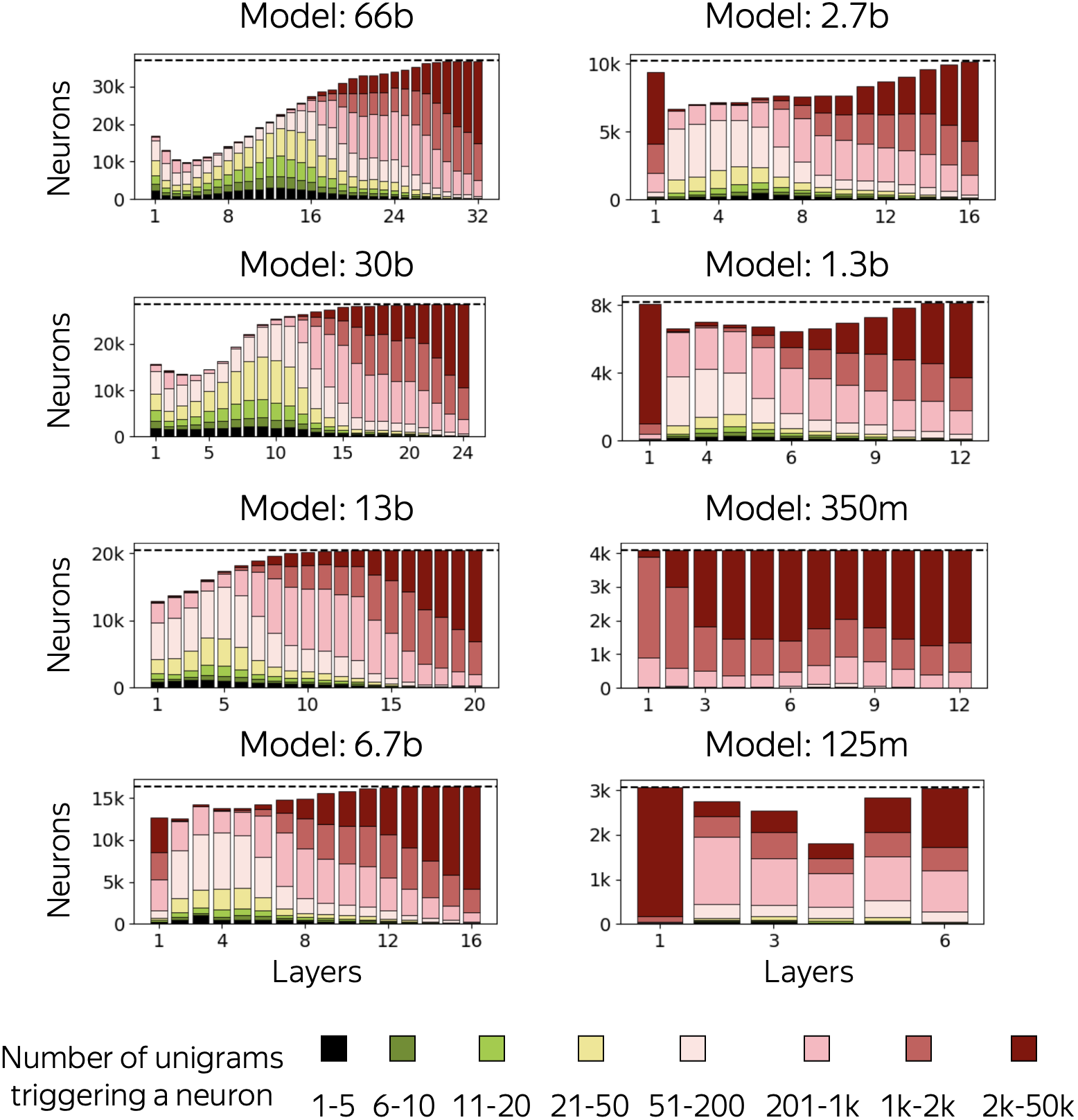}}
\caption{Neurons categorized by the number of unigrams (i.e., tokens) able to trigger them. First half of the network, alive neurons only.}
\label{fig:unigrams_covering_neuron_hist}
\vspace{-2ex}
\end{figure}

\section{N-gram-Detecting Neurons}
\label{sect:n_gram_neurons_full_section}

Now, let us look more closely into the patterns encoded in the lower half of the models and try to understand the nature of the observed above sparsity. Specifically, we analyze how neuron activations depend on an input n-gram.
For each input text with tokens $x_1, x_2, ..., x_S$, we record neuron activations at each position and if a neuron is activated (i.e., non-zero) at position $k$, we say that the n-gram $(x_{k-n+1}, \dots, x_k)$ \textit{triggered} this neuron.

In Sections~\ref{sect:n_grams_triggering_neuron}-\ref{sect:n_gram_neuron_suppress_their_triggers} we talk about unigrams (i.e., tokens) and come to larger n-grams in Section~\ref{sect:n_gram_neurons_beyond_unigrams}.

\subsection{Number of N-grams Triggering a Neuron}
\label{sect:n_grams_triggering_neuron}

First, let us see how many n-grams are able to trigger each neuron. For each neuron we evaluate the number of n-grams that cover at least $95\%$ of the neuron's activations. For the bottom half of the network, Figure~\ref{fig:unigrams_covering_neuron_hist} shows how neurons in each layer are categorized by the number of covering them n-grams (we show unigrams here and larger n-grams in Appendix~\ref{sect_apx:ngram_neurons}).

We see that, as anticipated, neurons in larger models are covered by less n-grams. Also, the largest models have a substantial proportion of neurons that are covered by as few as 1 to 5 tokens. This agrees with our hypothesis in the previous section: the model spreads discreet shallow patterns across specifically dedicated neurons.\footnote{Note that the 350m model does not follow the same pattern as all the rest: we will discuss this model in Section~\ref{sect:350m}.}

\begin{figure}[t!]
    \centering
    \begin{subfigure}[b]{0.24\textwidth}
        \includegraphics[width=\textwidth]{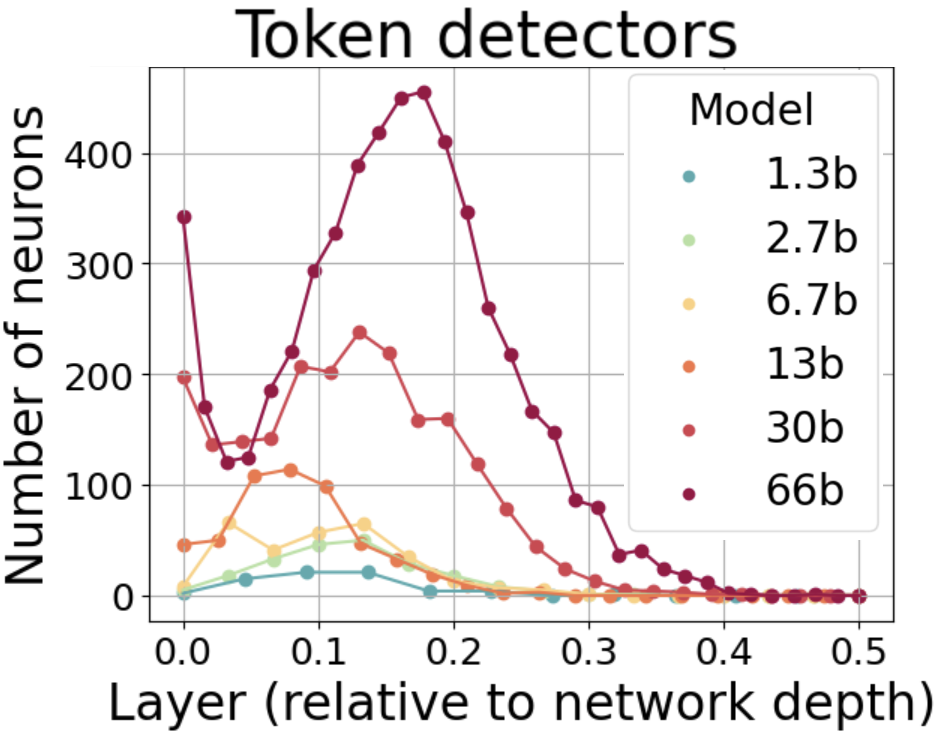}
        \caption{}
        \label{fig:ngram_neurons}
    \end{subfigure}
    \ \ 
    \begin{subfigure}[b]{0.22\textwidth}
         \includegraphics[width=\textwidth]{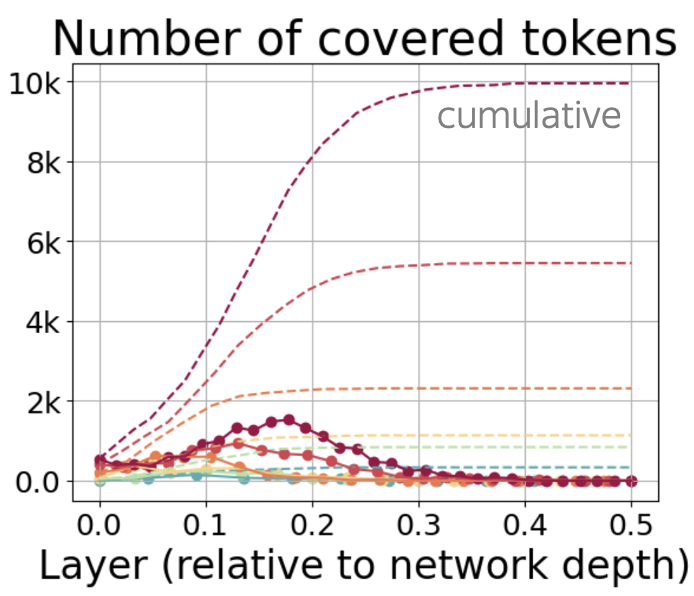}
        \caption{}
        \label{fig:number_of_covered_tokens}
    \end{subfigure}
    \vspace{-1ex}
    \caption{(a)~Number of token-detecting neurons; (b)~number of tokens that have a detecting them neuron: solid line~-- per layer, dashed~-- cumulative over layers.}
    \label{fig:ngram_neurons_both}
\end{figure}

\subsection{Token-Detecting Neurons}
\label{sect:n_gram_detectors_main_graphs}
Presence of neurons that can be triggered by only a few  (e.g., 1-5) tokens point to the possibility that some neurons act as token detectors, i.e. activate if and only if the input is one of the corresponding tokens, regardless of the previous context. To find such neurons, we (1)~pick neurons that can be triggered by only 1-5 tokens, (2)~gather tokens that are \textit{covered} by this neuron (if the neuron activates at least $95\%$ of the time the token is present), (3)~if altogether, these covered tokens are responsible for at least $95\%$ of neuron activations.\footnote{We exclude the begin-of-sentence token from these computations because for many neurons, this token is responsible for the majority of the activations.} 

\begin{figure*}[t]
\centering
{\includegraphics[scale=0.27]{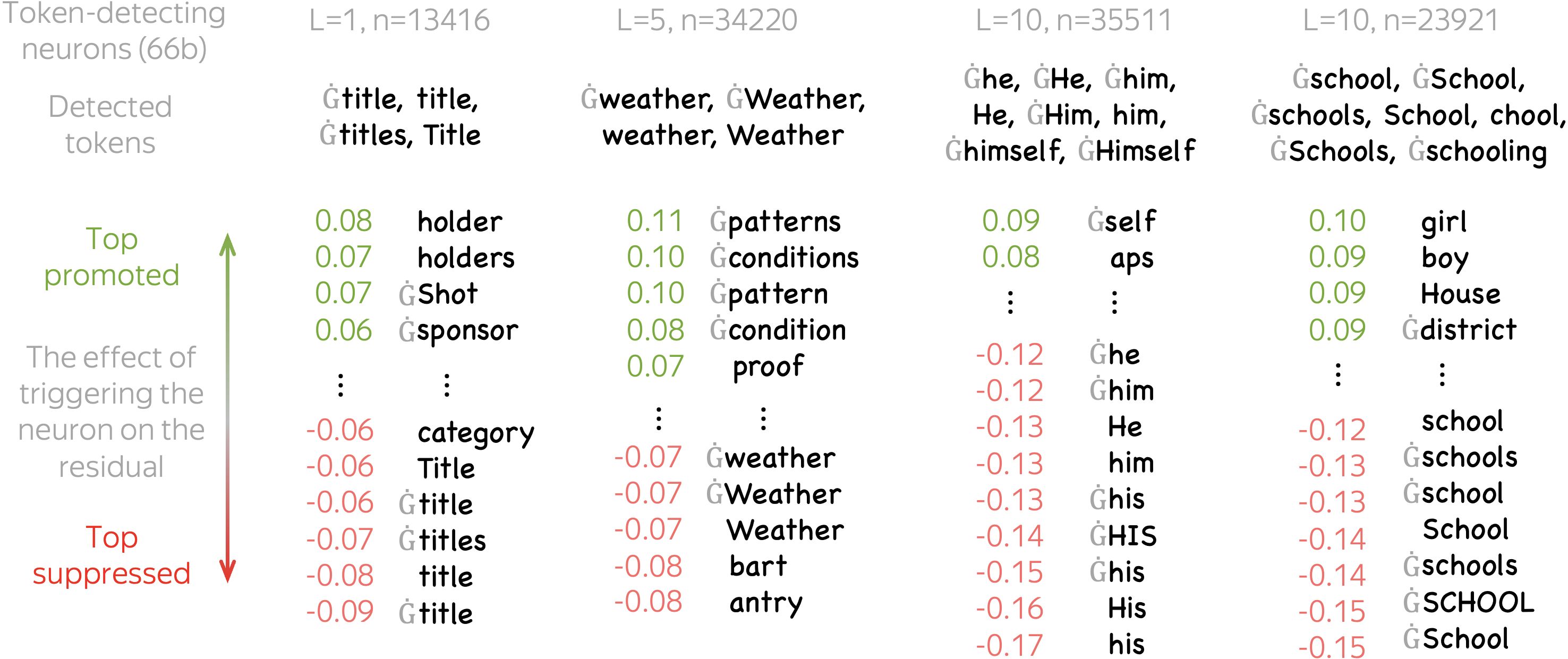}}
\caption{Examples of the top promoted and suppressed tokens for token-detecting neurons (\textcolor{gray}{Ġ} is a special symbol denoting the space before word~-- in the OPT tokenizers, it is part of a word); OPT-66b model.} 
\label{fig:ffn_concepts_examples}
\vspace{-2ex}
\end{figure*}

Figure~\ref{fig:ngram_neurons} shows that there are indeed a lot of token-detecting neurons. As expected, larger models have more such neurons and the 66b model has overall 5351 token detectors. Note that each token detector is responsible for a group of several tokens that, in most of the cases, are variants of the same word (e.g., with differences only in capitalization, presence of the space-before-word special symbol, morphological form, etc.). Figure~\ref{fig:ffn_concepts_examples} (top) shows examples of groups of tokens detected by token-detecting neurons.

Interestingly, the behavior of the largest models (starting from 13b of parameters) differs from that of the rest. While for smaller models the number of token detectors increases then goes down, larger models operate in three monotonic stages and start having many token-detecting neurons from the very first layer (Figures~\ref{fig:ngram_neurons_both}). This already shows qualitative differences between the models: with more capacity, larger models perform more complicated reasoning with more distinct stages.

\begin{figure}[t]
\centering
{\includegraphics[scale=0.27]{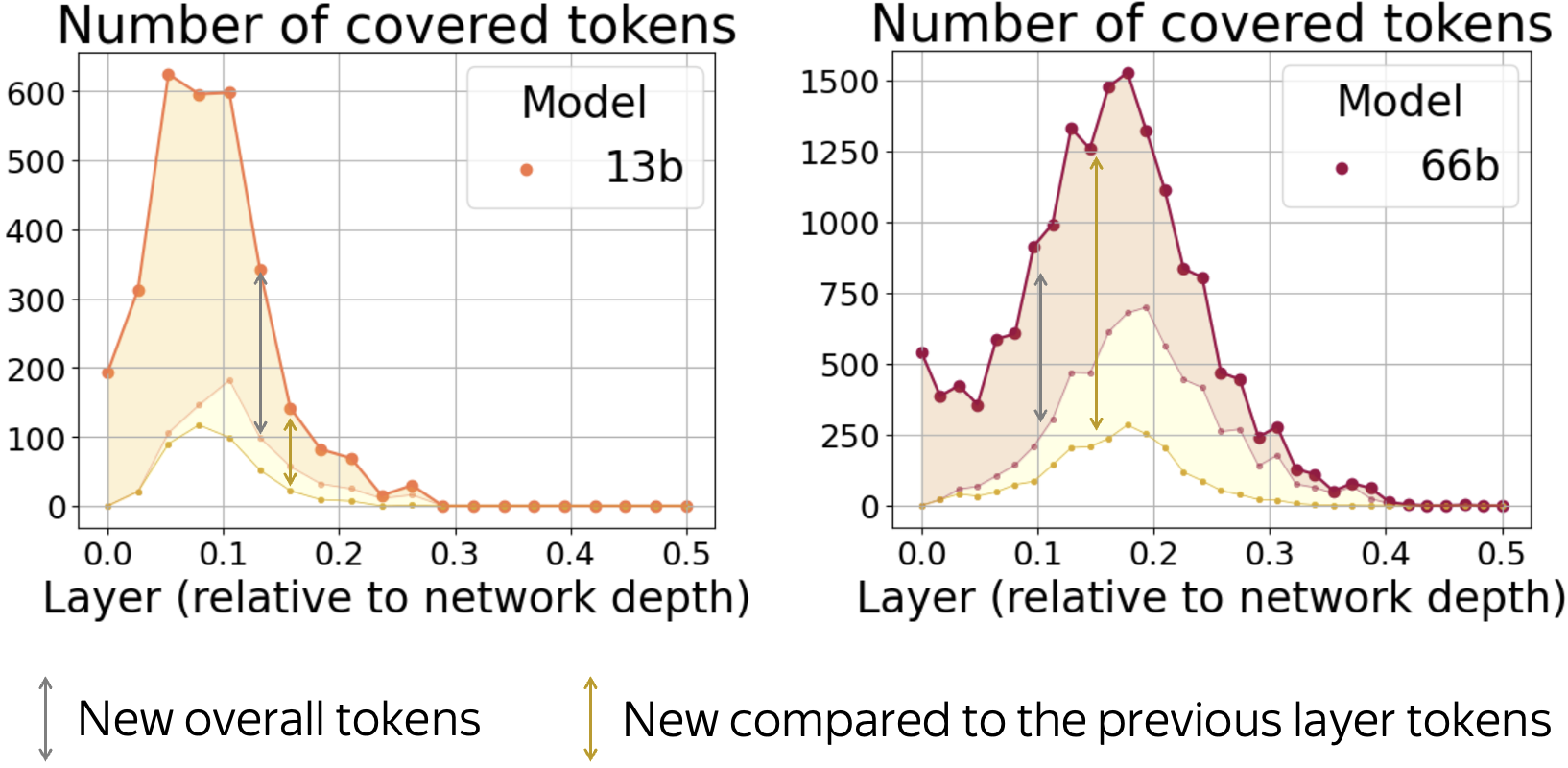}}
\caption{Number of tokens covered in each layer with indicated (i) new overall, and (ii) new compared to the previous layer tokens.}
\label{fig:covered_tokens_with_new}
\end{figure}

\subsection{Ensemble-Like Behaviour of the Layers}

Now, let us look at ``detected'' tokens, i.e. tokens that have a specialized detecting them neuron. Figure~\ref{fig:number_of_covered_tokens} shows the number of detected tokens in each layer as well as cumulative over layers number of detected tokens. We see that, e.g., the 66b model focuses on no more than 1.5k tokens in each layer but over 10k tokens overall. This means that across layers, token-detecting neurons are responsible for largely differing tokens. Indeed, Figure~\ref{fig:covered_tokens_with_new} shows that in each following layer, detected tokens mostly differ from all the tokens covered by the layers below.
All in all, this points to an ensemble-like (as opposed to sequential) behavior of the layers: layers collaborate so that token-detecting neurons cover largely different tokens in different layers. This divide-and-conquer-style strategy allows larger models to cover many tokens overall and use their capacity more effectively.

Originally, such an ensemble-like behavior of deep residual networks was observed in computer vision models~\cite{veit2016_resnet_as_ensemble}. For transformers, previous evidence includes simple experiments showing that e.g. dropping or reordering layers does not influence performance much~\cite{Fan2020LayerDrop,zhao2021nonlinearity}.

\begin{figure}[t]
\centering
{\includegraphics[scale=0.27]{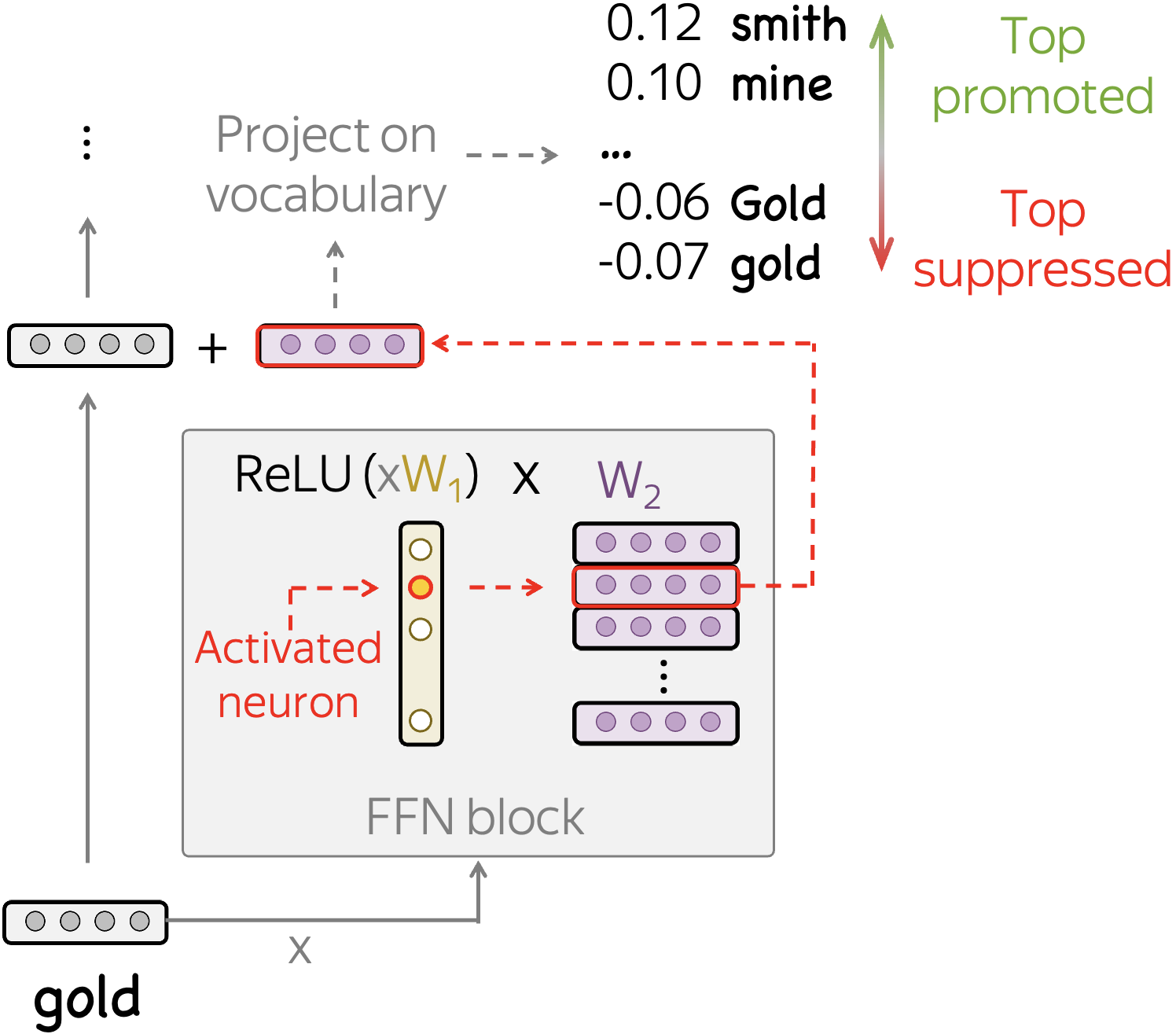}}
\caption{Intuition behind concept suppression:
we look not only at the top projections of an FFN update on vocabulary but also at the bottom. The concepts that are added with a negative value are suppressed.}
\label{fig:suppressed_concepts}
\vspace{-2ex}
\end{figure}

\subsection{Token Detectors Suppress Their Triggers}
\label{sect:n_gram_neuron_suppress_their_triggers}

Now let us try to understand the role of token-detecting neurons in the model by interpreting how they update the residual stream. Throughout the layers, token representation in the residual stream gets transformed from the token embedding for the current input token\footnote{For OPT models, along with an absolute positional embedding. 
} to the representation that encodes a distribution for the next token. 
This transformation happens via additive updates coming from attention and FFN blocks in each layer.
Whenever an FFN neuron is activated, the corresponding row of the second FFN layer (multiplied by this neuron's value) is added to the residual stream (see illustration in Figure~\ref{fig:suppressed_concepts}). By projecting this FFN row onto vocabulary, we can get an interpretation of this update (and, thus, the role of this neuron) in terms of its influence on the output distribution encoded in the residual stream.

\paragraph{Current token suppression: implicit or explicit?} Previously, this influence was understood only in terms of the top projections, i.e. tokens that are promoted~\cite{geva-etal-2021-transformer,geva-etal-2022-transformer}. This reflects an existing view supporting implicit rather than explicit loss of the current token identity over the course of layers. Namely, the view that the current identity gets ``buried'' as a result of updates useful for the next token as opposed to being removed by the model explicitly. In contrast, we look not only at the top projections but also at the bottom: if these projections are negative, the corresponding tokens are suppressed by the model (Figure~\ref{fig:suppressed_concepts}).

\begin{figure*}[t]
\centering
{\includegraphics[scale=0.27]{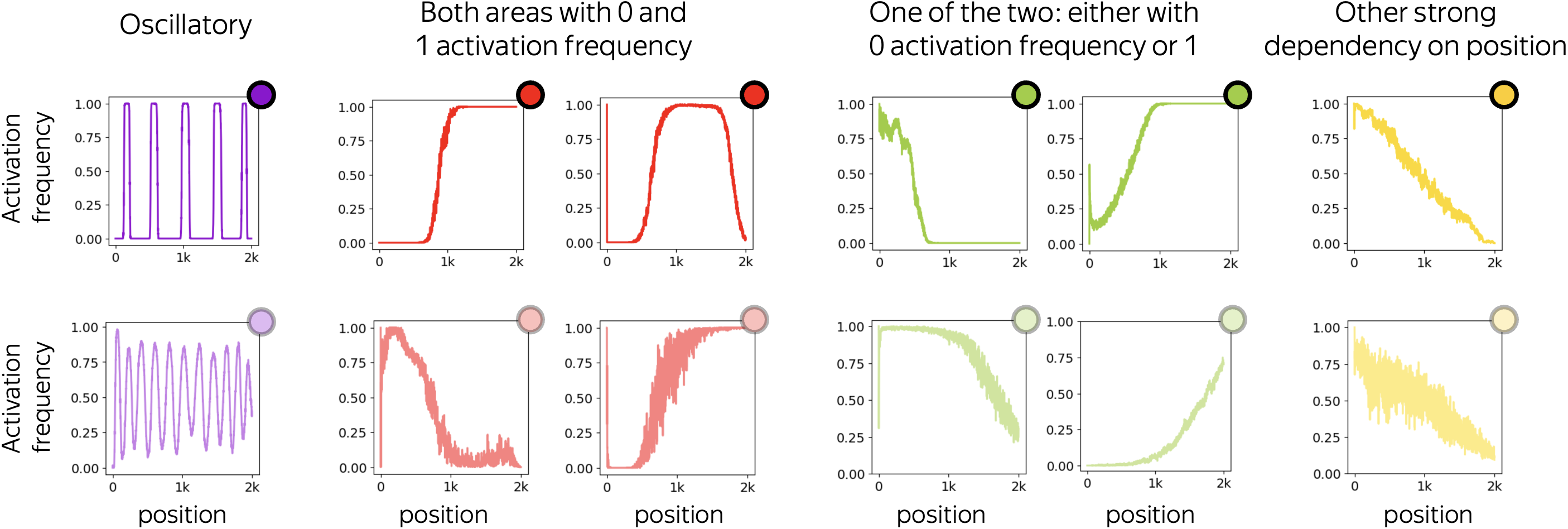}}
\caption{Types of positional neurons. Top row -- “strong” pattern, bottom row -- “weak” pattern.}
\label{fig:positional_types}
\vspace{-2ex}
\end{figure*}

\paragraph{Explicit token suppression in the model.}
 We find that often token-detecting neurons \textit{deliberately suppress the tokens they detect}. Figure~\ref{fig:ffn_concepts_examples} shows several examples of token-detecting neurons along with the top promoted and suppressed concepts. While the top promoted concepts are in line with previous work (they are potential next token candidates which agrees with~\citet{geva-etal-2021-transformer,geva-etal-2022-transformer}), the top suppressed concepts are rather unexpected: they are exactly the tokens triggering this neuron. This means that vector updates corresponding to these neurons point in the direction of the next token candidates at the same time as they point away from the tokens triggering the neuron. Note that this is not trivial since  these updates play two very different roles at the same time. Overall, for over $80\%$ of token-detecting neurons their corresponding updates point in the negative direction from the triggering them tokens (although, the triggering tokens are not always at the very top suppressed concepts as in the examples in Figure~\ref{fig:suppressed_concepts}).

Overall, we argue that models can have mechanisms that are targeted at removing information from the residual stream which can be explored further in future work.

\subsection{Beyond Unigrams}
\label{sect:n_gram_neurons_beyond_unigrams}

In Appendix~\ref{sect_apx:ngram_neurons}, we  show results for bigrams and trigrams that mirror our observations for unigrams: (i) larger models have more specialized neurons, (ii)~in each layer, models cover mostly new n-grams. Interestingly, for larger n-grams we see a more drastic gap between larger and smaller models.

\section{Positional Neurons}
\label{sect:positional}

When analyzing dead neurons (Section~\ref{sect:dead_neurons}), we also noticed some neurons that, consistently across diverse data, never activate except for a few first token positions. This motivates us to look further into how position is encoded in the model  and, specifically, whether some neurons are responsible for encoding positional information.

\subsection{Identifying Positional Neurons}

Intuitively, we want to find neurons whose activation patterns are defined by or, at least, strongly depend on token position. Formally, we identify neurons whose activations have high mutual information with position. For each neuron, we evaluate mutual information between two random variables: 
\begin{itemize}
    \item $act$~-- neuron is activated or not ($\{Y, N\}$),
    \item $pos$~-- token position ($\{1, 2, \dots, T\}$).
\end{itemize}

\paragraph{Formal setting.} We gather neuron activations for full-length data (i.e., $T=2048$ tokens) for Wikipedia, DM Mathematics and Codeparrot. Let $fr^{(pos)}_{n}$ be activation frequency of neuron $n$ at position $pos$ and $fr_n$ be the total activation frequency of this neuron. Then the desired mutual information is as follows:\footnote{For more details, see appendix~\ref{sect_apdx:mutual_info}.}
\begin{align*}
I(act, pos)=\frac {1}{T}\cdot\sum\limits_{pos=1}^{T}  \biggl[fr^{(pos)}_{n} \cdot \log{\frac{fr^{(pos)}_{n}}{fr_{n}}}+\\ (1-fr^{(pos)}_{n}) \cdot \log{\frac{1-fr^{(pos)}_{n}}{1-fr_{n}}}\biggr].    
\end{align*}

\paragraph{Choosing the neurons.} We pick neurons with $I(act, pos) > 0.05$, i.e. high mutual information with position~-- this gives neurons whose activation frequency depends on position rather than content. Indeed, if e.g. a neuron is always activated within certain position range regardless of data domain, we can treat this neuron as responsible for position; at least, to a certain extent.

\begin{figure*}[t]
\centering
{\includegraphics[scale=0.35]{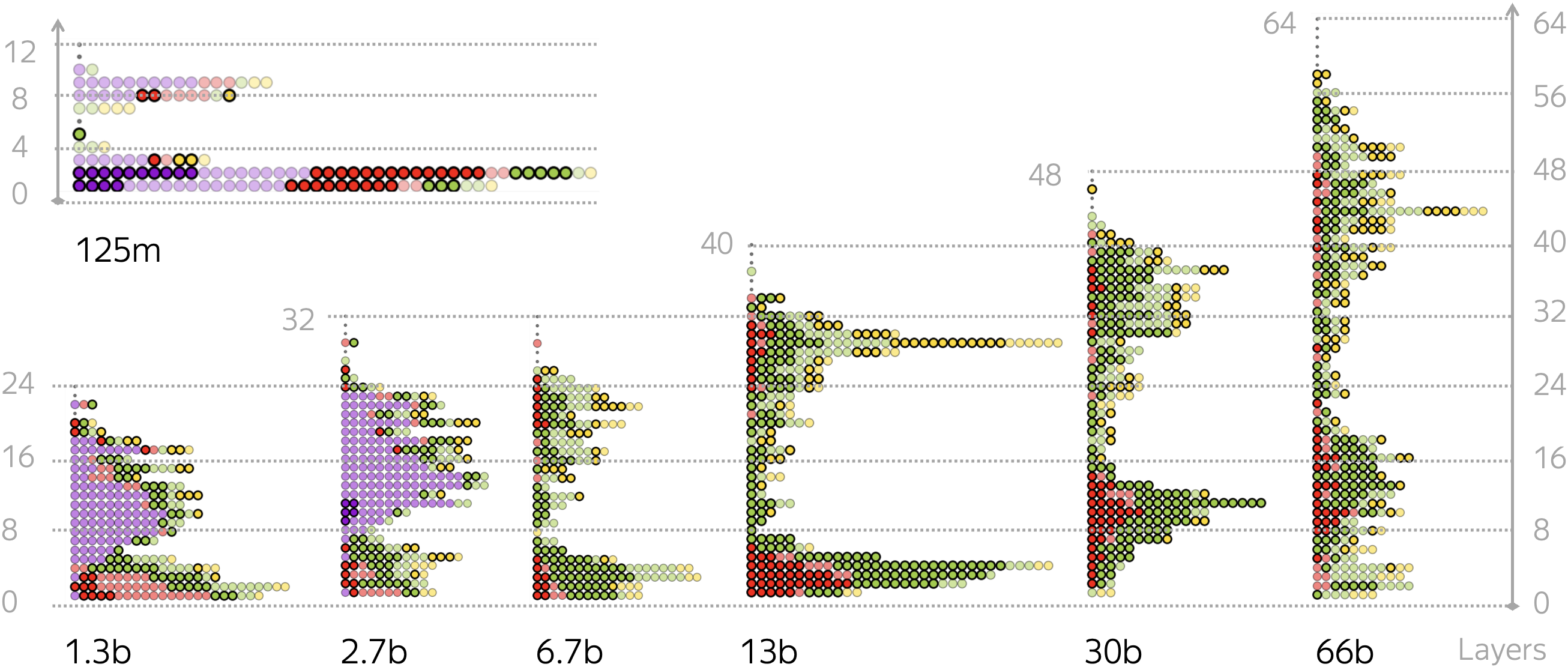}}
\caption{Positional neurons in each of the models. Each circle corresponds to a single neuron, colors and their intensity correspond to the types of patterns shown in Figure~\ref{fig:positional_types}.}
\label{fig:positional_neurons}
\end{figure*}

\subsection{Types of Positional Neurons}

After selecting positional neurons, we categorize them according to their activation pattern, i.e. activation frequency depending on position (Figure~\ref{fig:positional_types}).

\paragraph{Oscillatory.} These neurons are shown in purple in Figure~\ref{fig:positional_types}. When such a pattern is strong (top row), the activation pattern is an \textit{indicator function} of position ranges. In other words, such a neuron is activated if and only if the position falls into a certain set. Note that since the activation pattern does not change across data domains, it is defined solely by position and not the presence of some lexical or semantic information.

\paragraph{Both types of activation extremes.} These are the neurons whose activation pattern is not oscillatory but still has intervals where activation frequency reaches both ``activation extremes'': 0~(never activated) and 1~(always activated). Most frequently, such a neuron is activated only for positions less than or greater than some value and not activated otherwise. Similarly to oscillatory neurons, when such a pattern is strong (Figure~\ref{fig:positional_types}, top row), it is also (almost) an indicator function.

\paragraph{Only one type of activation extremes.} Differently from the previous two types, activation patterns for these neurons can reach only one of the extreme values 0 or 1 (Figure~\ref{fig:positional_types}, green). While this means that they never behave as indicator functions, there are position ranges where a neuron being activated or not depends solely on token position.

\paragraph{Other.} Finally, these are the neurons whose activation patterns strongly depend on position 
but do not have intervals where activation frequency stays 0 or 1 (Figure~\ref{fig:positional_types}, yellow). Typically, these activation patterns have lower mutual information with position than the previous three types.

\paragraph{Strong vs weak pattern.} We also distinguish ``strong'' and ``weak'' versions of each type which we will further denote with color intensity (Figure~\ref{fig:positional_types}, top vs bottom rows). For the first three types of positional neurons, the difference between strong and weak patterns lies in whether on the corresponding position ranges activation frequency equals 0 (or~1) or close, but not equals, to 0 (or~1). For the last type, this difference lies in how well we can predict activation frequency on a certain position knowing this value for the neighboring positions (informally, ``thin'' vs ``thick'' graph).

\subsection{Positional Neurons Across the Models}

For each of the models, Figure~\ref{fig:positional_neurons} illustrates the positional neurons across layers.

\paragraph{Small models encode position more explicitly.} First, we notice that smaller models rely substantially on oscillatory neurons: this is the most frequent type of positional neurons for models smaller than 6.7b of parameters. In combination with  many ``red'' neurons acting as indicator functions for wider position ranges, the model is able to derive token's absolute position rather accurately. Interestingly, larger models do not have oscillatory neurons and rely on more generic patterns shown with red- and green-colored circles. We can also see that from 13b to 66b, the model loses two-sided red neurons and uses the one-sided green ones more. This hints at one of the qualitative differences between smaller and larger models: while the former encode absolute position more accurately, the latter ones are likely to rely on something more meaningful than absolute position. This complements recent work showing that absolute position encoding is harmful for length generalization in reasoning tasks~\cite{kazemnejad2023impact}. Differently from their experiments with same model size but various positional encodings, we track changes with scale. We see that, despite all models being trained with absolute positional encodings, stronger models tend to abstract away from absolute position.

\begin{figure}[t]
\centering
{\includegraphics[scale=0.4]{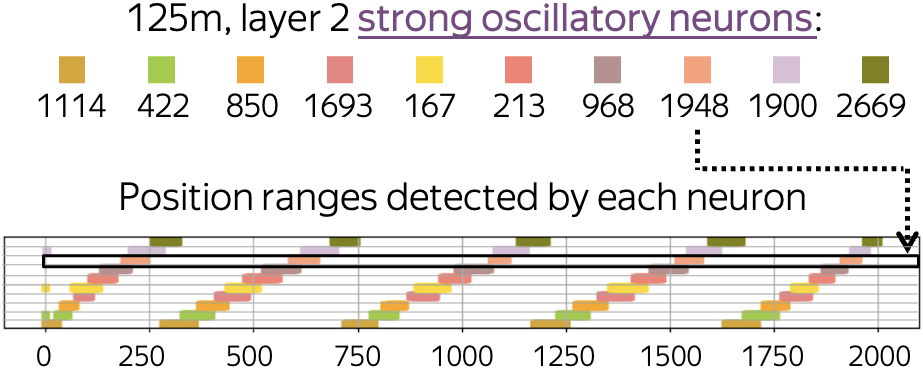}}
\caption{Position ranges indicated by strong oscillatory neurons in the second layer of the 125m model.}
\label{fig:pos_indcators_125m}
\end{figure}

\paragraph{Positional neurons work in teams.} Interestingly, positional neurons seem to collaborate to cover the full set of positions together. For example, let us look more closely at the 10 strongly oscillatory neurons in the second layer of the 125m model (shown with dark purple circles in Figure~\ref{fig:positional_neurons}). Since they act as indicator functions, we can plot position ranges indicated by each of these neurons.  Figure~\ref{fig:pos_indcators_125m} shows that (i)~indicated position ranges for these neurons are similar up to a shift, (ii) the shifts are organized in a ``perfect'' order in a sense that altogether, these ten neurons efficiently cover all positions such that none of these neurons is redundant.

\begin{figure*}[t]
\centering
{\includegraphics[scale=0.33]{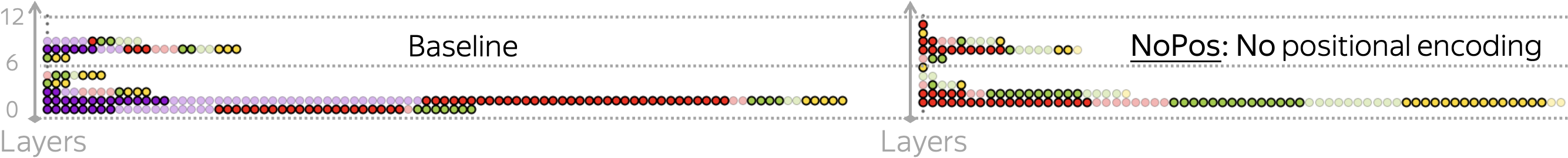}}
\caption{Positional neurons in 125m models: baseline vs model without positional encoding. Both models were trained for 300k batches.}
\label{fig:125m_pos_vs_no_pos}
\end{figure*}

\paragraph{The two stages within the model.} Finally, 
Figure~\ref{fig:positional_neurons} reveals two stages of up-and-downs of positional information within the model: roughly, the first third of the model and the rest. Interestingly, preferences in positional patterns also change between the stages: e.g., preference for ``red'' neurons changes to oscillatory purple patterns for the 1.3b and 2.7b models, and ``red'' patterns become less important in the upper stage for the 13b and 30b models. Note that the first third of the model corresponds to the sparse stage with the dead neurons and n-gram detectors (Sections~\ref{sect:dead_neurons}, \ref{sect:n_gram_neurons_full_section}). Therefore, we can hypothesize that in these two stages, 
positional information is first used locally to detect shallow patterns, and then more globally to use longer contexts and help encode semantic information. 

Previously, the distinct bottom-up stages of processing inside language models were observed in \citet{voita-etal-2019-bottom}. The authors explained that the way representations gain and lose information throughout the layers is defined by the training objective and why, among other things, positional information should (and does) get lost. This agrees with our results in this work: we can see that while there are many positional patterns in the second stage, they are weaker than in the first stage.

\begin{figure}[t]
\centering
{\includegraphics[scale=0.3]{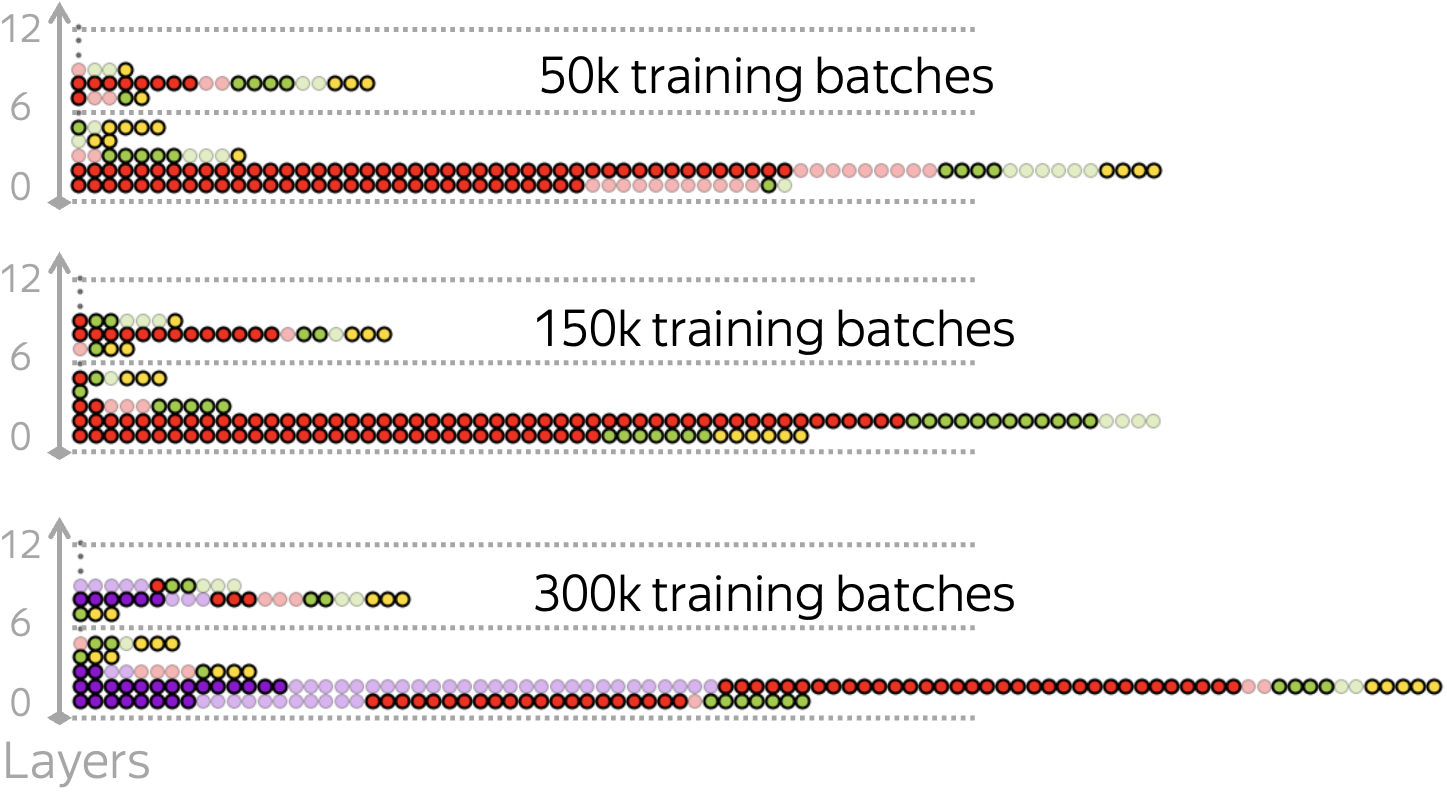}}
\caption{Positional neurons in the base 125m model trained with 50k, 150k and 300k batches.}
\label{fig:125m_small_training_steps}
\end{figure}

\subsection{Positional Neurons are Learned Even Without Positional Encoding}

Recently, it turned out that even without positional encoding, autoregressive language models still learn positional information~\cite{haviv-etal-2022-transformer}. We hypothesize that the mechanism these ``NoPos'' models use to encode position is positional neurons. To confirm this, we train two versions of the 125m model, with and without positional encodings, and compare the types of their positional neurons.

\paragraph{Setup.} We trained 125m models with the standard OPT setup but smaller training dataset: we used OpenWebText corpus~\cite{Gokaslan2019OpenWeb}, an open clone of the GPT-2 training data~\cite{radford2019language}. This dataset contains 3B tokens (compared 180B for OPT).

\paragraph{Positional neurons without positional encoding.} Figure~\ref{fig:125m_pos_vs_no_pos} shows positional neurons in two 125m models: trained with and without positional encoding.
We see that, indeed, the model without positional encoding also has many strong positional patterns. Note, however, that the NoPos model does not have oscillatory neurons which, in combination with other positional neurons, allow encoding absolute position rather accurately. This means that the NoPos model relies on more generic patterns, e.g. ``red'' neurons encoding whether a position is greater/less than some value.

\paragraph{Oscillatory neurons require longer training.} Finally, we found that oscillatory patterns appear only with long training. Figure~\ref{fig:125m_small_training_steps} shows positional patterns learned by the baseline 125m model trained for 50k, 150k and 300k training batches. We see that all models have very strong positional patterns, but only the last of them has oscillatory neurons. Apparently, learning absolute position requires longer training time.

\subsection{Doubting FFNs as Key-Value Memories}

Current widely held belief is that feed-forward layers in transformer-based language models operate as key-value memories. Specifically, ``each key correlates with textual patterns in the training examples, and each value induces a distribution over the output vocabulary''~(\citet{geva-etal-2021-transformer,geva-etal-2022-transformer,dai-etal-2022-knowledge,meng2022locating,ferrando-etal-2023-explaining}, among others). While in Section~\ref{sect:n_gram_neuron_suppress_their_triggers} we confirmed that this is true for some of the neurons, results in this section reveal that FFN layers can be used by the model in ways that \textit{do not fit the key-value memory view}. In particular, activations of strong positional neurons are defined by position regardless of textual content, and the corresponding values do not seem to encode meaningful distributions over vocabulary. This means that the role of these neurons is different from matching textual patterns to sets of the next token candidates. In a broader context, this means that the roles played by Transformer feed-forward layers are still poorly understood.

\section{The 350m Model: The Odd One Out}
\label{sect:350m}

As we already mentioned above, the 350m model does not follow the same pattern as the rest of the models. Specifically, it does not have dead neurons (Section~\ref{sect:dead_neurons}) and its neuron activations do not seem to be sparse with respect to triggering them n-grams as we saw for all the other models in Figure~\ref{fig:unigrams_covering_neuron_hist}.\footnote{There are, however, positional neurons; see Figure~\ref{fig:positional_neurons_350m} in Appendix~\ref{sect_apdx:350m_positional}).}

\paragraph{Modeling bits affect interpretability.} This becomes less surprizing when noticing that the 350m model is implemented differently from all the rest: it applies LayerNorm after attention and feed-forward blocks, while all the other models~-- before.\footnote{\url{https://github.com/huggingface/transformers/blob/main/src/transformers/models/opt/modeling_opt.py}} Apparently, such seemingly minor implementation details can affect interpretability of model components rather significantly. Indeed, previous work also tried choosing certain modeling aspects to encourage interpretability. Examples of such work include choosing an activation function to increase the number of interpretable neurons~\cite{elhage2022solu}, large body of work on sparse softmax variants to make output distributions or attention more interpretable~(\citet{sparsemax2016,Niculae_Blondel_sparse_attn_2017,peters-etal-2019-sparse,correia-etal-2019-adaptively,martins-etal-2020-sparse}, among others), or more extreme approaches with explicit modular structure that is aimed to be interpretable by construction~(\citet{Andreas_2016_CVPR,hu2019explainable,kirsch_2018_modular_networks,khot-etal-2021-text}, to name a few). Intuitively, choosing ReLU activation function as done in the OPT models can be seen as having the same motivation as developing sparse softmax variants: exact zeros in the model are inherently interpretable.

\section{Additional Related Work}

Historically, neurons have been a basic unit of analysis. Early works started from convolutional networks first for images~\cite{Krizhevsky_NIPS2012_image_patterns} and later for convolutional text classifiers~\cite{jacovi-etal-2018-understanding}. Similar to our work, \citet{jacovi-etal-2018-understanding} also find n-gram detectors; although, for small convolutional text classifiers this is an almost trivial observation compared to large Transformer-based language models as in our work. For recurrent networks, interpretable neurons include simple patterns such as line lengths, brackets and quotes~\cite{karpathy2015visualizing}, sentiment neuron~\cite{radford2017learning} and various neurons in machine translation models, such as tracking brackets, quotes, etc, as well as neurons correlated with higher-level concepts e.g. verb tense~\cite{bau2019neurons-in-mt}. For Transformer-based BERT, \citet{dai-etal-2022-knowledge} find that some neurons inside feed-forward blocks are responsible for storing factual knowledge.
Larger units of analysis include attention blocks~(\citet{voita-etal-2018-context,voita-etal-2019-analyzing,clark-etal-2019-bert,kovaleva-etal-2019-revealing,baan-2019-attention-in-summarization,correia-etal-2019-adaptively}, etc), feed-forward layers~\cite{geva-etal-2021-transformer,geva-etal-2022-transformer} and circuits responsible for certain tasks~\cite{wang2022interpretability,geva2023dissecting,hanna2023does}.

\section*{Acknowledgements}

The authors thank Nicola Cancedda, Yihong Chen, Igor Tufanov and FAIR London team for fruitful discussions and helpful feedback.

\bibliography{anthology,custom}

\begin{thebibliography}{51}
\expandafter\ifx\csname natexlab\endcsname\relax\def\natexlab#1{#1}\fi

\bibitem[{Andreas et~al.(2016)Andreas, Rohrbach, Darrell, and
  Klein}]{Andreas_2016_CVPR}
Jacob Andreas, Marcus Rohrbach, Trevor Darrell, and Dan Klein. 2016.
\newblock Neural module networks.
\newblock In \emph{Proceedings of the IEEE Conference on Computer Vision and
  Pattern Recognition (CVPR)}.

\bibitem[{Anil et~al.(2023)Anil, Dai, Firat, Johnson, Lepikhin, Passos,
  Shakeri, Taropa, Bailey, Chen, Chu, Clark, Shafey, Huang, Meier-Hellstern,
  Mishra, Moreira, Omernick, Robinson, Ruder, Tay, Xiao, Xu, Zhang, Abrego,
  Ahn, Austin, Barham, Botha, Bradbury, Brahma, Brooks, Catasta, Cheng, Cherry,
  Choquette-Choo, Chowdhery, Crepy, Dave, Dehghani, Dev, Devlin, Díaz, Du,
  Dyer, Feinberg, Feng, Fienber, Freitag, Garcia, Gehrmann, Gonzalez, Gur-Ari,
  Hand, Hashemi, Hou, Howland, Hu, Hui, Hurwitz, Isard, Ittycheriah, Jagielski,
  Jia, Kenealy, Krikun, Kudugunta, Lan, Lee, Lee, Li, Li, Li, Li, Li, Lim, Lin,
  Liu, Liu, Maggioni, Mahendru, Maynez, Misra, Moussalem, Nado, Nham, Ni,
  Nystrom, Parrish, Pellat, Polacek, Polozov, Pope, Qiao, Reif, Richter, Riley,
  Ros, Roy, Saeta, Samuel, Shelby, Slone, Smilkov, So, Sohn, Tokumine, Valter,
  Vasudevan, Vodrahalli, Wang, Wang, Wang, Wang, Wieting, Wu, Xu, Xu, Xue, Yin,
  Yu, Zhang, Zheng, Zheng, Zhou, Zhou, Petrov, and Wu}]{anil2023palm}
Rohan Anil, Andrew~M. Dai, Orhan Firat, Melvin Johnson, Dmitry Lepikhin,
  Alexandre Passos, Siamak Shakeri, Emanuel Taropa, Paige Bailey, Zhifeng Chen,
  Eric Chu, Jonathan~H. Clark, Laurent~El Shafey, Yanping Huang, Kathy
  Meier-Hellstern, Gaurav Mishra, Erica Moreira, Mark Omernick, Kevin Robinson,
  Sebastian Ruder, Yi~Tay, Kefan Xiao, Yuanzhong Xu, Yujing Zhang,
  Gustavo~Hernandez Abrego, Junwhan Ahn, Jacob Austin, Paul Barham, Jan Botha,
  James Bradbury, Siddhartha Brahma, Kevin Brooks, Michele Catasta, Yong Cheng,
  Colin Cherry, Christopher~A. Choquette-Choo, Aakanksha Chowdhery, Clément
  Crepy, Shachi Dave, Mostafa Dehghani, Sunipa Dev, Jacob Devlin, Mark Díaz,
  Nan Du, Ethan Dyer, Vlad Feinberg, Fangxiaoyu Feng, Vlad Fienber, Markus
  Freitag, Xavier Garcia, Sebastian Gehrmann, Lucas Gonzalez, Guy Gur-Ari,
  Steven Hand, Hadi Hashemi, Le~Hou, Joshua Howland, Andrea Hu, Jeffrey Hui,
  Jeremy Hurwitz, Michael Isard, Abe Ittycheriah, Matthew Jagielski, Wenhao
  Jia, Kathleen Kenealy, Maxim Krikun, Sneha Kudugunta, Chang Lan, Katherine
  Lee, Benjamin Lee, Eric Li, Music Li, Wei Li, YaGuang Li, Jian Li, Hyeontaek
  Lim, Hanzhao Lin, Zhongtao Liu, Frederick Liu, Marcello Maggioni, Aroma
  Mahendru, Joshua Maynez, Vedant Misra, Maysam Moussalem, Zachary Nado, John
  Nham, Eric Ni, Andrew Nystrom, Alicia Parrish, Marie Pellat, Martin Polacek,
  Alex Polozov, Reiner Pope, Siyuan Qiao, Emily Reif, Bryan Richter, Parker
  Riley, Alex~Castro Ros, Aurko Roy, Brennan Saeta, Rajkumar Samuel, Renee
  Shelby, Ambrose Slone, Daniel Smilkov, David~R. So, Daniel Sohn, Simon
  Tokumine, Dasha Valter, Vijay Vasudevan, Kiran Vodrahalli, Xuezhi Wang,
  Pidong Wang, Zirui Wang, Tao Wang, John Wieting, Yuhuai Wu, Kelvin Xu, Yunhan
  Xu, Linting Xue, Pengcheng Yin, Jiahui Yu, Qiao Zhang, Steven Zheng,
  Ce~Zheng, Weikang Zhou, Denny Zhou, Slav Petrov, and Yonghui Wu. 2023.
\newblock \href {http://arxiv.org/abs/2305.10403} {Palm 2 technical report}.

\bibitem[{Baan et~al.(2019)Baan, ter Hoeve, van~der Wees, Schuth, and
  de~Rijke}]{baan-2019-attention-in-summarization}
Joris Baan, Maartje ter Hoeve, Marlies van~der Wees, Anne Schuth, and Maarten
  de~Rijke. 2019.
\newblock \href {https://doi.org/10.48550/ARXIV.1911.03898} {Understanding
  multi-head attention in abstractive summarization}.

\bibitem[{Bau et~al.(2019)Bau, Belinkov, Sajjad, Durrani, Dalvi, and
  Glass}]{bau2019neurons-in-mt}
Anthony Bau, Yonatan Belinkov, Hassan Sajjad, Nadir Durrani, Fahim Dalvi, and
  James Glass. 2019.
\newblock \href {https://openreview.net/pdf?id=H1z-PsR5KX} {Identifying and
  controlling important neurons in neural machine translation}.
\newblock In \emph{International Conference on Learning Representations}, New
  Orleans.

\bibitem[{Baumgartner et~al.(2020)Baumgartner, Zannettou, Keegan, Squire, and
  Blackburn}]{baumgartner2020pushshift}
Jason Baumgartner, Savvas Zannettou, Brian Keegan, Megan Squire, and Jeremy
  Blackburn. 2020.
\newblock \href {http://arxiv.org/abs/2001.08435} {The pushshift reddit
  dataset}.

\bibitem[{Brown et~al.(2020)Brown, Mann, Ryder, Subbiah, Kaplan, Dhariwal,
  Neelakantan, Shyam, Sastry, Askell, Agarwal, Herbert-Voss, Krueger, Henighan,
  Child, Ramesh, Ziegler, Wu, Winter, Hesse, Chen, Sigler, Litwin, Gray, Chess,
  Clark, Berner, McCandlish, Radford, Sutskever, and Amodei}]{brown-gpt3}
Tom Brown, Benjamin Mann, Nick Ryder, Melanie Subbiah, Jared~D Kaplan, Prafulla
  Dhariwal, Arvind Neelakantan, Pranav Shyam, Girish Sastry, Amanda Askell,
  Sandhini Agarwal, Ariel Herbert-Voss, Gretchen Krueger, Tom Henighan, Rewon
  Child, Aditya Ramesh, Daniel Ziegler, Jeffrey Wu, Clemens Winter, Chris
  Hesse, Mark Chen, Eric Sigler, Mateusz Litwin, Scott Gray, Benjamin Chess,
  Jack Clark, Christopher Berner, Sam McCandlish, Alec Radford, Ilya Sutskever,
  and Dario Amodei. 2020.
\newblock \href
  {https://proceedings.neurips.cc/paper/2020/file/1457c0d6bfcb4967418bfb8ac142f64a-Paper.pdf}
  {Language models are few-shot learners}.
\newblock In \emph{Advances in Neural Information Processing Systems},
  volume~33, pages 1877--1901. Curran Associates, Inc.

\bibitem[{Clark et~al.(2019)Clark, Khandelwal, Levy, and
  Manning}]{clark-etal-2019-bert}
Kevin Clark, Urvashi Khandelwal, Omer Levy, and Christopher~D. Manning. 2019.
\newblock \href {https://doi.org/10.18653/v1/W19-4828} {What does {BERT} look
  at? an analysis of {BERT}{'}s attention}.
\newblock In \emph{Proceedings of the 2019 ACL Workshop BlackboxNLP: Analyzing
  and Interpreting Neural Networks for NLP}, pages 276--286, Florence, Italy.
  Association for Computational Linguistics.

\bibitem[{Correia et~al.(2019)Correia, Niculae, and
  Martins}]{correia-etal-2019-adaptively}
Gon{\c{c}}alo~M. Correia, Vlad Niculae, and Andr{\'e} F.~T. Martins. 2019.
\newblock \href {https://doi.org/10.18653/v1/D19-1223} {Adaptively sparse
  transformers}.
\newblock In \emph{Proceedings of the 2019 Conference on Empirical Methods in
  Natural Language Processing and the 9th International Joint Conference on
  Natural Language Processing (EMNLP-IJCNLP)}, pages 2174--2184, Hong Kong,
  China. Association for Computational Linguistics.

\bibitem[{Dai et~al.(2022)Dai, Dong, Hao, Sui, Chang, and
  Wei}]{dai-etal-2022-knowledge}
Damai Dai, Li~Dong, Yaru Hao, Zhifang Sui, Baobao Chang, and Furu Wei. 2022.
\newblock \href {https://doi.org/10.18653/v1/2022.acl-long.581} {Knowledge
  neurons in pretrained transformers}.
\newblock In \emph{Proceedings of the 60th Annual Meeting of the Association
  for Computational Linguistics (Volume 1: Long Papers)}, pages 8493--8502,
  Dublin, Ireland. Association for Computational Linguistics.

\bibitem[{Elhage et~al.(2022)Elhage, Hume, Olsson, Nanda, Henighan, Johnston,
  ElShowk, Joseph, DasSarma, Mann, Hernandez, Askell, Ndousse, Jones, , Drain,
  Chen, Bai, Ganguli, Lovitt, Hatfield-Dodds, Kernion, Conerly, Kravec, Fort,
  Kadavath, Jacobson, Tran-Johnson, Kaplan, Clark, Brown, McCandlish, Amodei,
  and Olah}]{elhage2022solu}
Nelson Elhage, Tristan Hume, Catherine Olsson, Neel Nanda, Tom Henighan, Scott
  Johnston, Sheer ElShowk, Nicholas Joseph, Nova DasSarma, Ben Mann, Danny
  Hernandez, Amanda Askell, Kamal Ndousse, Jones, , Dawn Drain, Anna Chen,
  Yuntao Bai, Deep Ganguli, Liane Lovitt, Zac Hatfield-Dodds, Jackson Kernion,
  Tom Conerly, Shauna Kravec, Stanislav Fort, Saurav Kadavath, Josh Jacobson,
  Eli Tran-Johnson, Jared Kaplan, Jack Clark, Tom Brown, Sam McCandlish, Dario
  Amodei, and Christopher Olah. 2022.
\newblock Softmax linear units.
\newblock Https://transformer-circuits.pub/2022/solu/index.html.

\bibitem[{Elhage et~al.(2021)Elhage, Nanda, Olsson, Henighan, Joseph, Mann,
  Askell, Bai, Chen, Conerly, DasSarma, Drain, Ganguli, Hatfield-Dodds,
  Hernandez, Jones, Kernion, Lovitt, Ndousse, Amodei, Brown, Clark, Kaplan,
  McCandlish, and Olah}]{elhage2021mathematical}
Nelson Elhage, Neel Nanda, Catherine Olsson, Tom Henighan, Nicholas Joseph, Ben
  Mann, Amanda Askell, Yuntao Bai, Anna Chen, Tom Conerly, Nova DasSarma, Dawn
  Drain, Deep Ganguli, Zac Hatfield-Dodds, Danny Hernandez, Andy Jones, Jackson
  Kernion, Liane Lovitt, Kamal Ndousse, Dario Amodei, Tom Brown, Jack Clark,
  Jared Kaplan, Sam McCandlish, and Chris Olah. 2021.
\newblock \href {https://transformer-circuits.pub/2021/framework/index.html} {A
  mathematical framework for transformer circuits}.
\newblock \emph{Transformer Circuits Thread}.

\bibitem[{Fan et~al.(2020)Fan, Grave, and Joulin}]{Fan2020LayerDrop}
Angela Fan, Edouard Grave, and Armand Joulin. 2020.
\newblock \href {https://openreview.net/forum?id=SylO2yStDr} {Reducing
  transformer depth on demand with structured dropout}.
\newblock In \emph{International Conference on Learning Representations}.

\bibitem[{Ferrando et~al.(2023)Ferrando, G{\'a}llego, Tsiamas, and
  Costa-juss{\`a}}]{ferrando-etal-2023-explaining}
Javier Ferrando, Gerard~I. G{\'a}llego, Ioannis Tsiamas, and Marta~R.
  Costa-juss{\`a}. 2023.
\newblock \href {https://doi.org/10.18653/v1/2023.acl-long.301} {Explaining how
  transformers use context to build predictions}.
\newblock In \emph{Proceedings of the 61st Annual Meeting of the Association
  for Computational Linguistics (Volume 1: Long Papers)}, pages 5486--5513,
  Toronto, Canada. Association for Computational Linguistics.

\bibitem[{Gao et~al.(2020)Gao, Biderman, Black, Golding, Hoppe, Foster, Phang,
  He, Thite, Nabeshima, Presser, and Leahy}]{gao2020pile}
Leo Gao, Stella Biderman, Sid Black, Laurence Golding, Travis Hoppe, Charles
  Foster, Jason Phang, Horace He, Anish Thite, Noa Nabeshima, Shawn Presser,
  and Connor Leahy. 2020.
\newblock \href {http://arxiv.org/abs/2101.00027} {The pile: An 800gb dataset
  of diverse text for language modeling}.

\bibitem[{Geva et~al.(2023)Geva, Bastings, Filippova, and
  Globerson}]{geva2023dissecting}
Mor Geva, Jasmijn Bastings, Katja Filippova, and Amir Globerson. 2023.
\newblock \href {http://arxiv.org/abs/2304.14767} {Dissecting recall of factual
  associations in auto-regressive language models}.

\bibitem[{Geva et~al.(2022)Geva, Caciularu, Wang, and
  Goldberg}]{geva-etal-2022-transformer}
Mor Geva, Avi Caciularu, Kevin Wang, and Yoav Goldberg. 2022.
\newblock \href {https://aclanthology.org/2022.emnlp-main.3} {Transformer
  feed-forward layers build predictions by promoting concepts in the vocabulary
  space}.
\newblock In \emph{Proceedings of the 2022 Conference on Empirical Methods in
  Natural Language Processing}, pages 30--45, Abu Dhabi, United Arab Emirates.
  Association for Computational Linguistics.

\bibitem[{Geva et~al.(2021)Geva, Schuster, Berant, and
  Levy}]{geva-etal-2021-transformer}
Mor Geva, Roei Schuster, Jonathan Berant, and Omer Levy. 2021.
\newblock \href {https://doi.org/10.18653/v1/2021.emnlp-main.446} {Transformer
  feed-forward layers are key-value memories}.
\newblock In \emph{Proceedings of the 2021 Conference on Empirical Methods in
  Natural Language Processing}, pages 5484--5495, Online and Punta Cana,
  Dominican Republic. Association for Computational Linguistics.

\bibitem[{Gokaslan and Cohen(2019)}]{Gokaslan2019OpenWeb}
Aaron Gokaslan and Vanya Cohen. 2019.
\newblock \href {http://Skylion007.github.io/OpenWebTextCorpus} {Openwebtext
  corpus}.

\bibitem[{Hanna et~al.(2023)Hanna, Liu, and Variengien}]{hanna2023does}
Michael Hanna, Ollie Liu, and Alexandre Variengien. 2023.
\newblock \href {http://arxiv.org/abs/2305.00586} {How does gpt-2 compute
  greater-than?: Interpreting mathematical abilities in a pre-trained language
  model}.

\bibitem[{Haviv et~al.(2022)Haviv, Ram, Press, Izsak, and
  Levy}]{haviv-etal-2022-transformer}
Adi Haviv, Ori Ram, Ofir Press, Peter Izsak, and Omer Levy. 2022.
\newblock \href {https://aclanthology.org/2022.findings-emnlp.99} {Transformer
  language models without positional encodings still learn positional
  information}.
\newblock In \emph{Findings of the Association for Computational Linguistics:
  EMNLP 2022}, pages 1382--1390, Abu Dhabi, United Arab Emirates. Association
  for Computational Linguistics.

\bibitem[{Hu et~al.(2018)Hu, Andreas, Darrell, and Saenko}]{hu2019explainable}
Ronghang Hu, Jacob Andreas, Trevor Darrell, and Kate Saenko. 2018.
\newblock Explainable neural computation via stack neural module networks.
\newblock In \emph{Proceedings of the European conference on computer vision
  (ECCV)}.

\bibitem[{Jacovi et~al.(2018)Jacovi, Sar~Shalom, and
  Goldberg}]{jacovi-etal-2018-understanding}
Alon Jacovi, Oren Sar~Shalom, and Yoav Goldberg. 2018.
\newblock \href {https://doi.org/10.18653/v1/W18-5408} {Understanding
  convolutional neural networks for text classification}.
\newblock In \emph{Proceedings of the 2018 {EMNLP} Workshop {B}lackbox{NLP}:
  Analyzing and Interpreting Neural Networks for {NLP}}, pages 56--65,
  Brussels, Belgium. Association for Computational Linguistics.

\bibitem[{Jawahar et~al.(2019)Jawahar, Sagot, and
  Seddah}]{jawahar-etal-2019-bert}
Ganesh Jawahar, Beno{\^\i}t Sagot, and Djam{\'e} Seddah. 2019.
\newblock \href {https://doi.org/10.18653/v1/P19-1356} {What does {BERT} learn
  about the structure of language?}
\newblock In \emph{Proceedings of the 57th Annual Meeting of the Association
  for Computational Linguistics}, pages 3651--3657, Florence, Italy.
  Association for Computational Linguistics.

\bibitem[{Kaplan et~al.(2020)Kaplan, McCandlish, Henighan, Brown, Chess, Child,
  Gray, Radford, Wu, and Amodei}]{kaplan2020scaling}
Jared Kaplan, Sam McCandlish, Tom Henighan, Tom~B. Brown, Benjamin Chess, Rewon
  Child, Scott Gray, Alec Radford, Jeffrey Wu, and Dario Amodei. 2020.
\newblock \href {http://arxiv.org/abs/2001.08361} {Scaling laws for neural
  language models}.

\bibitem[{Karpathy et~al.(2015)Karpathy, Johnson, and
  Fei-Fei}]{karpathy2015visualizing}
Andrej Karpathy, Justin Johnson, and Li~Fei-Fei. 2015.
\newblock \href {http://arxiv.org/abs/1506.02078} {Visualizing and
  understanding recurrent networks}.

\bibitem[{Kazemnejad et~al.(2023)Kazemnejad, Padhi, Ramamurthy, Das, and
  Reddy}]{kazemnejad2023impact}
Amirhossein Kazemnejad, Inkit Padhi, Karthikeyan~Natesan Ramamurthy, Payel Das,
  and Siva Reddy. 2023.
\newblock \href {http://arxiv.org/abs/2305.19466} {The impact of positional
  encoding on length generalization in transformers}.

\bibitem[{Khot et~al.(2021)Khot, Khashabi, Richardson, Clark, and
  Sabharwal}]{khot-etal-2021-text}
Tushar Khot, Daniel Khashabi, Kyle Richardson, Peter Clark, and Ashish
  Sabharwal. 2021.
\newblock \href {https://doi.org/10.18653/v1/2021.naacl-main.99} {Text modular
  networks: Learning to decompose tasks in the language of existing models}.
\newblock In \emph{Proceedings of the 2021 Conference of the North American
  Chapter of the Association for Computational Linguistics: Human Language
  Technologies}, pages 1264--1279, Online. Association for Computational
  Linguistics.

\bibitem[{Kirsch et~al.(2018)Kirsch, Kunze, and
  Barber}]{kirsch_2018_modular_networks}
Louis Kirsch, Julius Kunze, and David Barber. 2018.
\newblock \href
  {https://proceedings.neurips.cc/paper_files/paper/2018/file/310ce61c90f3a46e340ee8257bc70e93-Paper.pdf}
  {Modular networks: Learning to decompose neural computation}.
\newblock In \emph{Advances in Neural Information Processing Systems},
  volume~31. Curran Associates, Inc.

\bibitem[{Kovaleva et~al.(2019)Kovaleva, Romanov, Rogers, and
  Rumshisky}]{kovaleva-etal-2019-revealing}
Olga Kovaleva, Alexey Romanov, Anna Rogers, and Anna Rumshisky. 2019.
\newblock \href {https://doi.org/10.18653/v1/D19-1445} {Revealing the dark
  secrets of {BERT}}.
\newblock In \emph{Proceedings of the 2019 Conference on Empirical Methods in
  Natural Language Processing and the 9th International Joint Conference on
  Natural Language Processing (EMNLP-IJCNLP)}, pages 4365--4374, Hong Kong,
  China. Association for Computational Linguistics.

\bibitem[{Krizhevsky et~al.(2012)Krizhevsky, Sutskever, and
  Hinton}]{Krizhevsky_NIPS2012_image_patterns}
Alex Krizhevsky, Ilya Sutskever, and Geoffrey~E Hinton. 2012.
\newblock \href
  {https://proceedings.neurips.cc/paper_files/paper/2012/file/c399862d3b9d6b76c8436e924a68c45b-Paper.pdf}
  {Imagenet classification with deep convolutional neural networks}.
\newblock In \emph{Advances in Neural Information Processing Systems},
  volume~25. Curran Associates, Inc.

\bibitem[{Liu et~al.(2019)Liu, Gardner, Belinkov, Peters, and
  Smith}]{liu-etal-2019-linguistic}
Nelson~F. Liu, Matt Gardner, Yonatan Belinkov, Matthew~E. Peters, and Noah~A.
  Smith. 2019.
\newblock \href {https://doi.org/10.18653/v1/N19-1112} {Linguistic knowledge
  and transferability of contextual representations}.
\newblock In \emph{Proceedings of the 2019 Conference of the North {A}merican
  Chapter of the Association for Computational Linguistics: Human Language
  Technologies, Volume 1 (Long and Short Papers)}, pages 1073--1094,
  Minneapolis, Minnesota. Association for Computational Linguistics.

\bibitem[{Martins and Astudillo(2016)}]{sparsemax2016}
Andr\'{e} F.~T. Martins and Ram\'{o}n~F. Astudillo. 2016.
\newblock From softmax to sparsemax: A sparse model of attention and
  multi-label classification.
\newblock In \emph{Proceedings of the 33rd International Conference on
  International Conference on Machine Learning - Volume 48}, ICML'16, page
  1614–1623. JMLR.org.

\bibitem[{Martins et~al.(2020)Martins, Marinho, and
  Martins}]{martins-etal-2020-sparse}
Pedro~Henrique Martins, Zita Marinho, and Andr{\'e} F.~T. Martins. 2020.
\newblock \href {https://doi.org/10.18653/v1/2020.emnlp-main.348} {Sparse text
  generation}.
\newblock In \emph{Proceedings of the 2020 Conference on Empirical Methods in
  Natural Language Processing (EMNLP)}, pages 4252--4273, Online. Association
  for Computational Linguistics.

\bibitem[{Meng et~al.(2022)Meng, Bau, Andonian, and
  Belinkov}]{meng2022locating}
Kevin Meng, David Bau, Alex~J Andonian, and Yonatan Belinkov. 2022.
\newblock \href {https://openreview.net/forum?id=-h6WAS6eE4} {Locating and
  editing factual associations in {GPT}}.
\newblock In \emph{Advances in Neural Information Processing Systems}.

\bibitem[{Niculae and Blondel(2017)}]{Niculae_Blondel_sparse_attn_2017}
Vlad Niculae and Mathieu Blondel. 2017.
\newblock \href
  {https://proceedings.neurips.cc/paper_files/paper/2017/file/2d1b2a5ff364606ff041650887723470-Paper.pdf}
  {A regularized framework for sparse and structured neural attention}.
\newblock In \emph{Advances in Neural Information Processing Systems},
  volume~30. Curran Associates, Inc.

\bibitem[{OpenAI(2023)}]{openai2023gpt4}
OpenAI. 2023.
\newblock \href {http://arxiv.org/abs/2303.08774} {Gpt-4 technical report}.

\bibitem[{Ouyang et~al.(2022)Ouyang, Wu, Jiang, Almeida, Wainwright, Mishkin,
  Zhang, Agarwal, Slama, Ray, Schulman, Hilton, Kelton, Miller, Simens, Askell,
  Welinder, Christiano, Leike, and Lowe}]{ouyang2022training}
Long Ouyang, Jeff Wu, Xu~Jiang, Diogo Almeida, Carroll~L. Wainwright, Pamela
  Mishkin, Chong Zhang, Sandhini Agarwal, Katarina Slama, Alex Ray, John
  Schulman, Jacob Hilton, Fraser Kelton, Luke Miller, Maddie Simens, Amanda
  Askell, Peter Welinder, Paul Christiano, Jan Leike, and Ryan Lowe. 2022.
\newblock \href {http://arxiv.org/abs/2203.02155} {Training language models to
  follow instructions with human feedback}.

\bibitem[{Peters et~al.(2019)Peters, Niculae, and
  Martins}]{peters-etal-2019-sparse}
Ben Peters, Vlad Niculae, and Andr{\'e} F.~T. Martins. 2019.
\newblock \href {https://doi.org/10.18653/v1/P19-1146} {Sparse
  sequence-to-sequence models}.
\newblock In \emph{Proceedings of the 57th Annual Meeting of the Association
  for Computational Linguistics}, pages 1504--1519, Florence, Italy.
  Association for Computational Linguistics.

\bibitem[{Peters et~al.(2018)Peters, Neumann, Iyyer, Gardner, Clark, Lee, and
  Zettlemoyer}]{peters-etal-2018-deep}
Matthew~E. Peters, Mark Neumann, Mohit Iyyer, Matt Gardner, Christopher Clark,
  Kenton Lee, and Luke Zettlemoyer. 2018.
\newblock \href {https://doi.org/10.18653/v1/N18-1202} {Deep contextualized
  word representations}.
\newblock In \emph{Proceedings of the 2018 Conference of the North {A}merican
  Chapter of the Association for Computational Linguistics: Human Language
  Technologies, Volume 1 (Long Papers)}, pages 2227--2237, New Orleans,
  Louisiana. Association for Computational Linguistics.

\bibitem[{Radford et~al.(2017)Radford, Jozefowicz, and
  Sutskever}]{radford2017learning}
Alec Radford, Rafal Jozefowicz, and Ilya Sutskever. 2017.
\newblock \href {http://arxiv.org/abs/1704.01444} {Learning to generate reviews
  and discovering sentiment}.

\bibitem[{Radford et~al.(2019)Radford, Wu, Child, Luan, Amodei, and
  Sutskever}]{radford2019language}
Alec Radford, Jeffrey Wu, Rewon Child, David Luan, Dario Amodei, and Ilya
  Sutskever. 2019.
\newblock Language models are unsupervised multitask learners.
\newblock \emph{OpenAI Blog}, 1(8):9.

\bibitem[{Roller et~al.(2021)Roller, Dinan, Goyal, Ju, Williamson, Liu, Xu,
  Ott, Smith, Boureau, and Weston}]{roller-etal-2021-recipes}
Stephen Roller, Emily Dinan, Naman Goyal, Da~Ju, Mary Williamson, Yinhan Liu,
  Jing Xu, Myle Ott, Eric~Michael Smith, Y-Lan Boureau, and Jason Weston. 2021.
\newblock \href {https://doi.org/10.18653/v1/2021.eacl-main.24} {Recipes for
  building an open-domain chatbot}.
\newblock In \emph{Proceedings of the 16th Conference of the European Chapter
  of the Association for Computational Linguistics: Main Volume}, pages
  300--325, Online. Association for Computational Linguistics.

\bibitem[{Tenney et~al.(2019)Tenney, Das, and Pavlick}]{tenney-etal-2019-bert}
Ian Tenney, Dipanjan Das, and Ellie Pavlick. 2019.
\newblock \href {https://doi.org/10.18653/v1/P19-1452} {{BERT} rediscovers the
  classical {NLP} pipeline}.
\newblock In \emph{Proceedings of the 57th Annual Meeting of the Association
  for Computational Linguistics}, pages 4593--4601, Florence, Italy.
  Association for Computational Linguistics.

\bibitem[{Veit et~al.(2016)Veit, Wilber, and
  Belongie}]{veit2016_resnet_as_ensemble}
Andreas Veit, Michael~J Wilber, and Serge Belongie. 2016.
\newblock \href
  {https://proceedings.neurips.cc/paper_files/paper/2016/file/37bc2f75bf1bcfe8450a1a41c200364c-Paper.pdf}
  {Residual networks behave like ensembles of relatively shallow networks}.
\newblock In \emph{Advances in Neural Information Processing Systems},
  volume~29. Curran Associates, Inc.

\bibitem[{Voita et~al.(2019{\natexlab{a}})Voita, Sennrich, and
  Titov}]{voita-etal-2019-bottom}
Elena Voita, Rico Sennrich, and Ivan Titov. 2019{\natexlab{a}}.
\newblock \href {https://doi.org/10.18653/v1/D19-1448} {The bottom-up evolution
  of representations in the transformer: A study with machine translation and
  language modeling objectives}.
\newblock In \emph{Proceedings of the 2019 Conference on Empirical Methods in
  Natural Language Processing and the 9th International Joint Conference on
  Natural Language Processing (EMNLP-IJCNLP)}, pages 4396--4406, Hong Kong,
  China. Association for Computational Linguistics.

\bibitem[{Voita et~al.(2018)Voita, Serdyukov, Sennrich, and
  Titov}]{voita-etal-2018-context}
Elena Voita, Pavel Serdyukov, Rico Sennrich, and Ivan Titov. 2018.
\newblock \href {https://doi.org/10.18653/v1/P18-1117} {Context-aware neural
  machine translation learns anaphora resolution}.
\newblock In \emph{Proceedings of the 56th Annual Meeting of the Association
  for Computational Linguistics (Volume 1: Long Papers)}, pages 1264--1274,
  Melbourne, Australia. Association for Computational Linguistics.

\bibitem[{Voita et~al.(2019{\natexlab{b}})Voita, Talbot, Moiseev, Sennrich, and
  Titov}]{voita-etal-2019-analyzing}
Elena Voita, David Talbot, Fedor Moiseev, Rico Sennrich, and Ivan Titov.
  2019{\natexlab{b}}.
\newblock \href {https://doi.org/10.18653/v1/P19-1580} {Analyzing multi-head
  self-attention: Specialized heads do the heavy lifting, the rest can be
  pruned}.
\newblock In \emph{Proceedings of the 57th Annual Meeting of the Association
  for Computational Linguistics}, pages 5797--5808, Florence, Italy.
  Association for Computational Linguistics.

\bibitem[{Wang et~al.(2022)Wang, Variengien, Conmy, Shlegeris, and
  Steinhardt}]{wang2022interpretability}
Kevin Wang, Alexandre Variengien, Arthur Conmy, Buck Shlegeris, and Jacob
  Steinhardt. 2022.
\newblock \href {http://arxiv.org/abs/2211.00593} {Interpretability in the
  wild: a circuit for indirect object identification in gpt-2 small}.

\bibitem[{Wei et~al.(2022)Wei, Tay, Bommasani, Raffel, Zoph, Borgeaud,
  Yogatama, Bosma, Zhou, Metzler, Chi, Hashimoto, Vinyals, Liang, Dean, and
  Fedus}]{wei2022emergent}
Jason Wei, Yi~Tay, Rishi Bommasani, Colin Raffel, Barret Zoph, Sebastian
  Borgeaud, Dani Yogatama, Maarten Bosma, Denny Zhou, Donald Metzler, Ed~H.
  Chi, Tatsunori Hashimoto, Oriol Vinyals, Percy Liang, Jeff Dean, and William
  Fedus. 2022.
\newblock \href {https://openreview.net/forum?id=yzkSU5zdwD} {Emergent
  abilities of large language models}.
\newblock \emph{Transactions on Machine Learning Research}.
\newblock Survey Certification.

\bibitem[{Zhang et~al.(2022)Zhang, Roller, Goyal, Artetxe, Chen, Chen, Dewan,
  Diab, Li, Lin, Mihaylov, Ott, Shleifer, Shuster, Simig, Koura, Sridhar, Wang,
  and Zettlemoyer}]{zhang2022opt}
Susan Zhang, Stephen Roller, Naman Goyal, Mikel Artetxe, Moya Chen, Shuohui
  Chen, Christopher Dewan, Mona Diab, Xian Li, Xi~Victoria Lin, Todor Mihaylov,
  Myle Ott, Sam Shleifer, Kurt Shuster, Daniel Simig, Punit~Singh Koura, Anjali
  Sridhar, Tianlu Wang, and Luke Zettlemoyer. 2022.
\newblock \href {http://arxiv.org/abs/2205.01068} {Opt: Open pre-trained
  transformer language models}.

\bibitem[{Zhao et~al.(2021)Zhao, Pascual, Brunner, and
  Wattenhofer}]{zhao2021nonlinearity}
Sumu Zhao, Damian Pascual, Gino Brunner, and Roger Wattenhofer. 2021.
\newblock \href {http://arxiv.org/abs/2101.04547} {Of non-linearity and
  commutativity in bert}.

\end{thebibliography}
\bibliographystyle{acl_natbib}

\newpage
\appendix

\section{N-gram-Detecting Neurons}
\label{sect_apx:ngram_neurons}

\subsection{Number of N-grams Triggering a Neuron}

\begin{figure}[t]
\centering
{\includegraphics[scale=0.27]{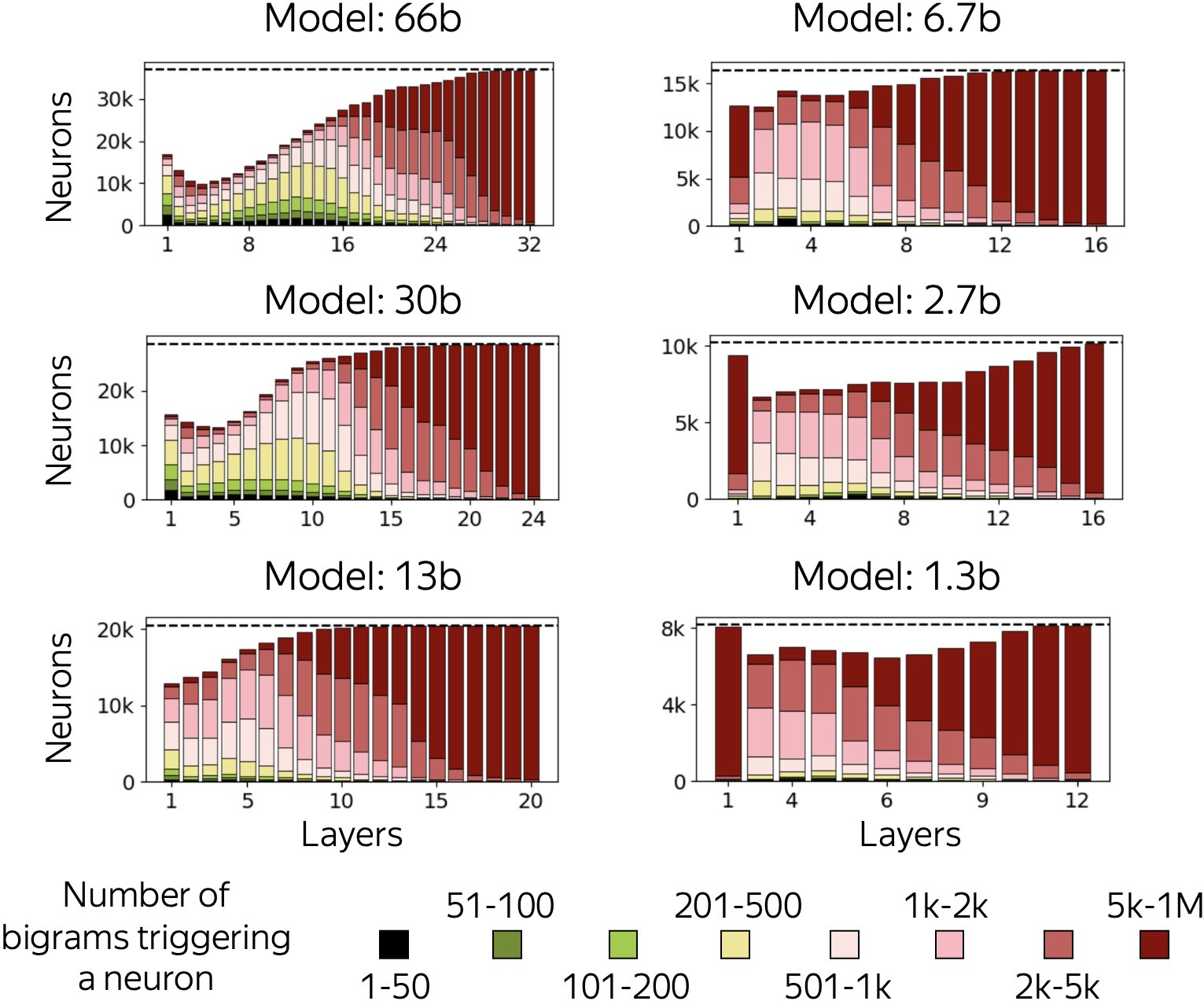}}
\caption{Neurons categorized by the number of bigrams able to trigger them. First half of the network, alive neurons only.}
\label{fig:bigrams_covering_neuron_hist}
\end{figure}

\begin{figure}[t]
\centering
{\includegraphics[scale=0.27]{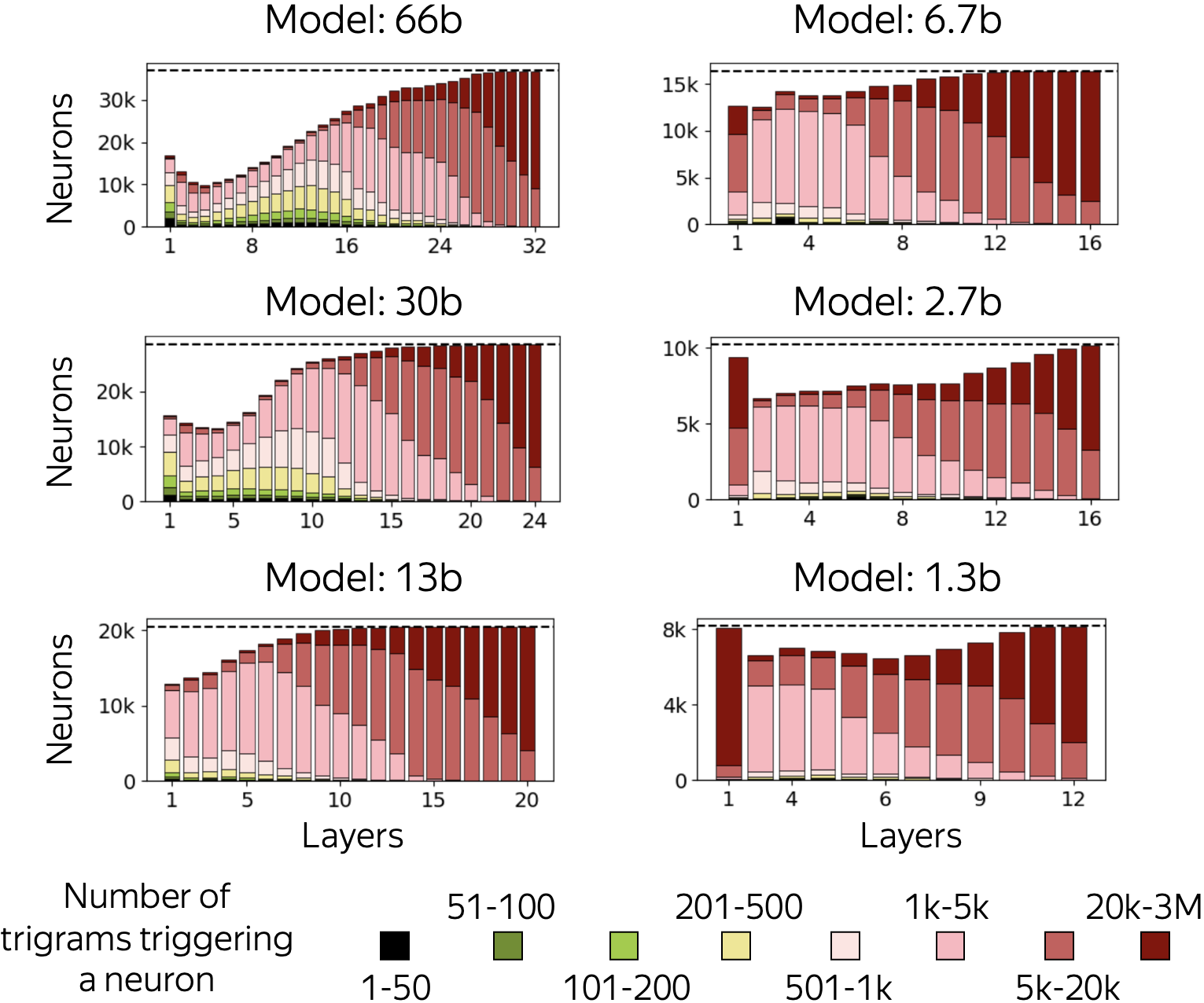}}
\caption{Neurons categorized by the number of trigrams able to trigger them. First half of the network, alive neurons only.}
\label{fig:trigrams_covering_neuron_hist}
\end{figure}

Figure~\ref{fig:bigrams_covering_neuron_hist} shows how neurons in each layer are categorized by the number of covering them bigrams, Figure~\ref{fig:trigrams_covering_neuron_hist}~-- trigrams. As expected, neurons in larger models are covered by less n-grams.

\subsection{Trigram-Detecting Neurons}

Similarly to token-detecting neurons in Section~\ref{sect:n_gram_detectors_main_graphs}, we also find neurons that are specialized on 3-grams. Specifically, we (1) pick neurons that are covered by only 1-50 trigrams, (2) gather trigrams that are covered by this neuron (if the neuron activated at least $95\%$ of the time the trigram is present), (3) if altogether, these covered trigrams are responsible for at least $95\%$ of neuron activations.

\begin{figure}[t!]
    \centering
    \begin{subfigure}[b]{0.24\textwidth}
        \includegraphics[width=\textwidth]{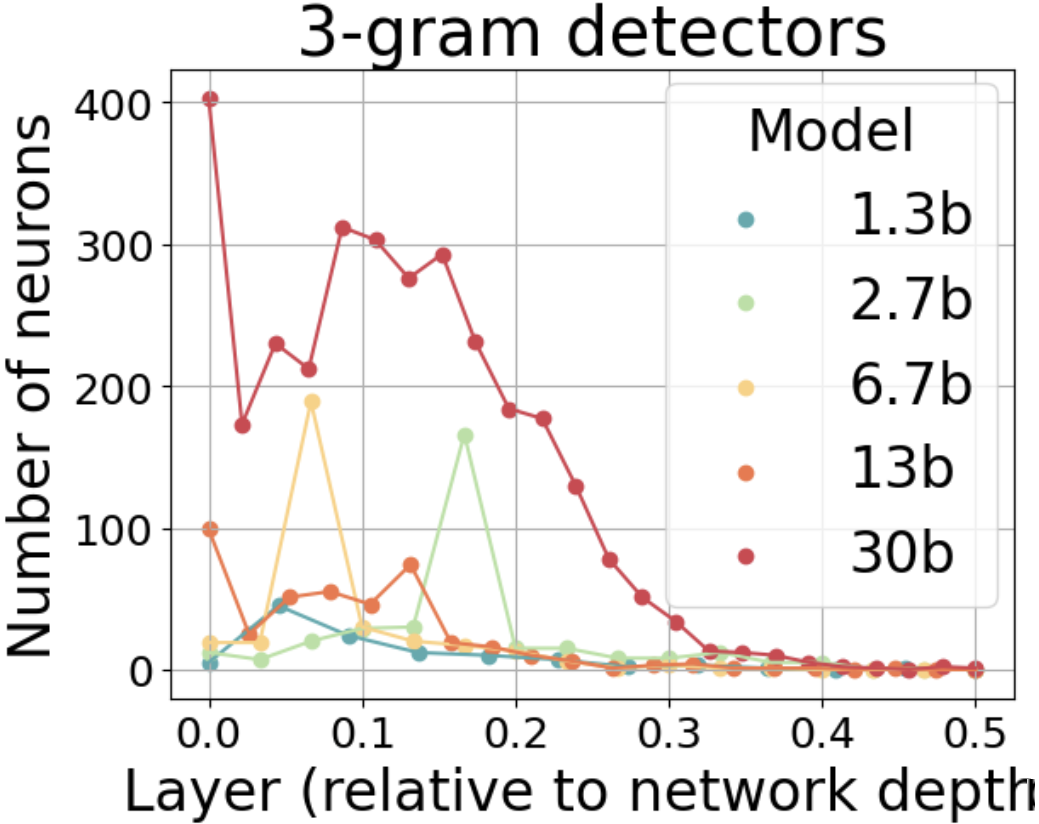}
        \caption{}
        \label{fig:trigram_neurons}
    \end{subfigure}
    \ \ 
    \begin{subfigure}[b]{0.22\textwidth}
         \includegraphics[width=\textwidth]{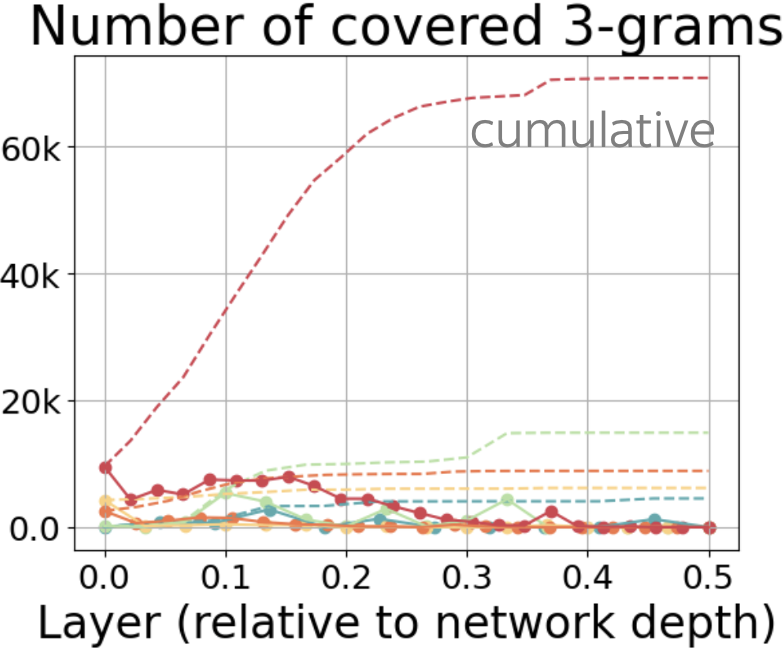}
        \caption{}
        \label{fig:number_of_covered_trigrams}
    \end{subfigure}
    \vspace{-1ex}
    \caption{(a)~Number of trigram-detecting neurons; (b)~number of trigrams that have a detecting them neuron: solid line~-- per layer, dashed~-- cumulative over layers.}
    \label{fig:trigram_neurons_both}
\end{figure}

Figure~\ref{fig:trigram_neurons_both} shows the results. Overall, the results further support our main observations: larger models have more neurons responsible for n-grams. Interestingly, when looking at trigrams rather than tokens, at 30b of parameters we see a drastic jump in the number of covered n-grams. This indicates that
one of the qualitative differences between larger and smaller models lies in the expansion of the families of features they are able to represent.

\subsection{Ensemble-Like Layer Behavior}

\begin{figure}[t]
\centering
{\includegraphics[scale=0.27]{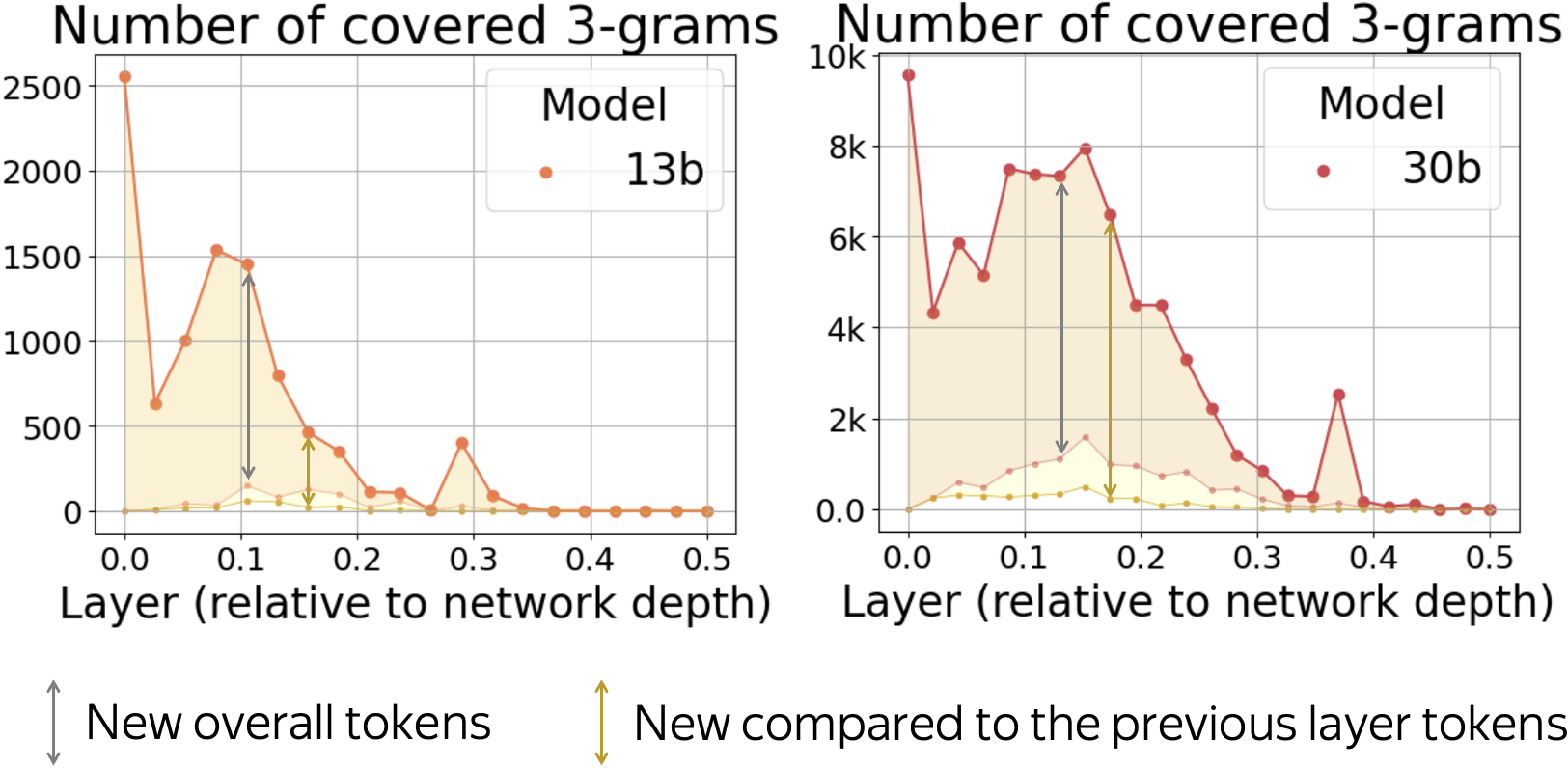}}
\caption{Number of trigrams covered in each layer with indicated (i) new overall, and (ii) new compared to the previous layer tokens.}
\label{fig:covered_trigrams_with_new}
\end{figure}

Figure~\ref{fig:covered_trigrams_with_new} shows the number of covered trigrams in each layer. We see that in each layer, models cover largely new trigrams.

\section{Positional Neurons}

\subsection{Mutual Information}
\label{sect_apdx:mutual_info}

For each neuron, we evaluate mutual information between two random variables: 
\begin{itemize}
    \item $act$~-- neuron is activated or not ($\{Y, N\}$),
    \item $pos$~-- token position ($\{1, 2, \dots, T\}$).
\end{itemize}

\paragraph{Formal setting.} We gather neuron activations for full-length data (i.e., $T=2048$ tokens) for Wikipedia, DM Mathematics and Codeparrot. Let $fr^{(pos)}_{n}$ be activation frequency of neuron $n$ at position $pos$ and $fr_n$ be the total activation frequency of this neuron. 

Then the desired mutual information is as follows:
$$I(act, pos) =$$ $$= \sum\limits_{act}\sum\limits_{pos=1}^{T} \frac {1}{p(pos)} p(act|pos) \cdot \log{\frac{p(act|pos)}{p(act)}}=
$$
Since we only feed full-length texts, all positions appear with the same frequency: $p(pos)=1/T$.

$$=\frac {1}{T}\cdot\sum\limits_{act \in \{Y, N\}}\sum\limits_{pos=1}^{T}  p(act|pos) \cdot \log{\frac{p(act|pos)}{p(act)}}=$$

$$=\frac{1}{T}\cdot\sum\limits_{pos=1}^{T}  p(act=Y|pos) \cdot \log{\frac{p(act=Y|pos)}{p(act=Y)}}+
$$

$$
\frac{1}{T}\cdot\!\!\sum\limits_{pos=1}^{T}  \!\!\left(1\!-\!p(act=Y|pos)\right) \cdot \log{\frac{1\!-\!p(act\!=\!Y|pos)}{1-p(act=Y)}}=$$

\begin{align*}
=\frac {1}{T}\cdot\sum\limits_{pos=1}^{T}  \biggl[fr^{(pos)}_{n} \cdot \log{\frac{fr^{(pos)}_{n}}{fr_{n}}}+\\ (1-fr^{(pos)}_{n}) \cdot \log{\frac{1-fr^{(pos)}_{n}}{1-fr_{n}}}\biggr].    
\end{align*}

\subsection{Positional Neurons for the 350m Model}
\label{sect_apdx:350m_positional}

\begin{figure}[t]
\centering
{\includegraphics[scale=0.45]{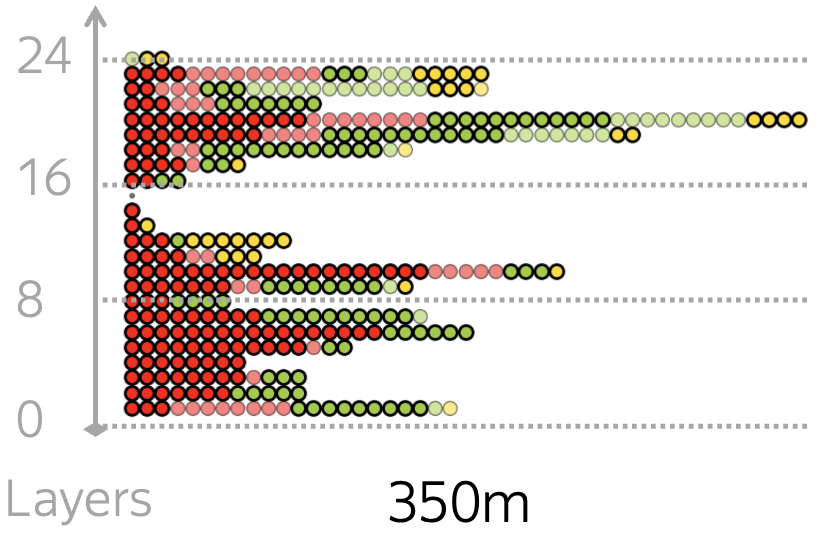}}
\caption{Positional neurons in the 350m model. Each circle corresponds to a single neuron, colors and their intensity correspond to the types of patterns shown in Figure~\ref{fig:positional_types}.}
\label{fig:positional_neurons_350m}
\end{figure}

The results are shown in Figure~\ref{fig:positional_neurons_350m}.

\end{document}